\documentclass[10pt,journal,compsoc]{IEEEtran}

\usepackage[nocompress]{cite}

\usepackage[utf8]{inputenc} % allow utf-8 input
\usepackage[T1]{fontenc}    % use 8-bit T1 fonts
\usepackage{hyperref}       % hyperlinks
\usepackage{url}            % simple URL typesetting
\usepackage{booktabs}       % professional-quality tables
\usepackage{amsfonts}       % blackboard math symbols
\usepackage{nicefrac}       % compact symbols for 1/2, etc.
\usepackage{microtype}      % microtypography
\usepackage{xcolor}         % colors
\usepackage{graphicx}
\usepackage{multirow}
\usepackage{algorithm}
\usepackage{algpseudocode}
\usepackage{wrapfig}
\usepackage{xr}
\usepackage{subcaption}
\usepackage{fontawesome5}
\usepackage{color,soul}
\usepackage{pifont}
\usepackage{tcolorbox}
\usepackage{tikz}
\usetikzlibrary{shapes, backgrounds, shadows}
\usepackage{ifthen}

\soulregister{\cite}{7}
\soulregister{\ref}{7}

\usepackage{amsmath,amsfonts,bm}

\def\1{\bm{1}}

\DeclareMathAlphabet{\mathsfit}{\encodingdefault}{\sfdefault}{m}{sl}
\SetMathAlphabet{\mathsfit}{bold}{\encodingdefault}{\sfdefault}{bx}{n}

\def\gP{{\mathcal{P}}}

\def\gX{{\mathcal{X}}}
\def\gY{{\mathcal{Y}}}

\def\sR{{\mathbb{R}}}

\newcommand{\Ls}{\mathcal{L}}

\DeclareMathOperator*{\argmin}{arg\,min}

\newcommand{\ex}{(x, y)}

\newcommand{\exn}[1]{(x_{#1}, y_{#1})}

\newcommand{\trainset}{\mathcal{D}}
\newcommand{\trainsetdef}{\{\exn{n}\}_{n=1}^N}
\newcommand{\trainstream}{\vec{\mathcal{D}}}
\newcommand{\trainstreamdef}{(\exn{t})_{t=1}^T}
\newcommand{\testset}{\mathcal{E}}
\newcommand{\testsetdef}{\{\exn{\tilde n}\}_{\tilde n = N+1}^{N + \tilde N}}
\newcommand{\epi}{(\trainset, \testset)}
\newcommand{\epim}[1]{(\trainset^{#1}, \testset^{#1})}

\newcommand{\mtrainset}{\mathbb{D}}
\newcommand{\mtrainsetdef}{\{\epim{m}\}_{m=1}^{M}}
\newcommand{\mtestset}{\mathbb{E}}
\newcommand{\mtestsetdef}{\{(\trainset^{\tilde m}, \testset^{\tilde m})\}_{\tilde m = 1}^{\tilde M}}

\newcommand{\mtrainstream}{\vec{\mathbb{D}}}
\newcommand{\mtrainstreamdef}{(\epim{u})_{u=1}^U}

\newcommand{\taskset}{\mathcal{T}}

\newcommand{\model}{f}
\newcommand{\modelparam}{\theta}
\newcommand{\learner}{G}
\newcommand{\learnerparam}{\omega}
\newcommand{\mlearner}{\mathrm{MetaUpdate}}
\newcommand{\loss}{\ell}

\definecolor{magentaX}{HTML}{F53D7A}

\makeatletter
\def\@IEEEsectpunct{~~}
\def\paragraph{\@startsection{paragraph}{4}{\z@}{1.5ex plus 1.5ex minus 0.5ex}%
{0ex}{\normalfont\normalsize\bfseries}}  % add \sffamily for san serif font
\makeatother

\begin{document}

\title{When Meta-Learning Meets Online and Continual Learning: A Survey}

\author{Jaehyeon~Son$^*$,
        Soochan~Lee$^*$,
        and~Gunhee~Kim % <-this % stops a space

\thanks{$*$ Equal contribution}
\thanks{J. Son and G. Kim are with Seoul National University, Seoul, Republic of Korea. S. Lee is with LG AI Research, Seoul, Republic of Korea. Gunhee Kim is the corresponding author.}
\thanks{E-mail: sjh9876@snu.ac.kr, soochan.lee@lgresearch.ai, gunhee@snu.ac.kr}
}

\IEEEtitleabstractindextext{%
\begin{abstract}
Over the past decade, deep neural networks have demonstrated significant success using the training scheme that involves mini-batch stochastic gradient descent on extensive datasets.
Expanding upon this accomplishment, there has been a surge in research exploring the application of neural networks in other learning scenarios.
One notable framework that has garnered significant attention is \emph{meta-learning}.
Often described as ``learning to learn,'' meta-learning is a data-driven approach to optimize the learning algorithm.
Other branches of interest are \emph{continual learning} and \emph{online learning}, both of which involve incrementally updating a model with streaming data.
While these frameworks were initially developed independently, recent works have started investigating their combinations, proposing novel problem settings and learning algorithms.
However, due to the elevated complexity and lack of unified terminology, discerning differences between the learning frameworks can be challenging even for experienced researchers.
To facilitate a clear understanding, this paper provides a comprehensive survey that organizes various problem settings using consistent terminology and formal descriptions.
By offering an overview of these learning paradigms, our work aims to foster further advancements in this promising area of research.
\end{abstract}

}

\maketitle

\section{Introduction}
\label{sec:introduction}

\IEEEPARstart{T}{he} recent success of deep neural networks is mostly based on the offline learning framework that involves stochastic gradient descent (SGD) with mini-batches sampled in an independent and identically distributed (i.i.d.) fashion from a large dataset.
On the other hand, humans demonstrate an impressive ability to learn incrementally, even from highly non-stationary data streams, and do not require an extensive amount of training data like deep neural networks.
Moreover, humans can continuously improve their learning capability as they accumulate more knowledge and experience.
As a result, there has been significant effort devoted to adapting deep neural networks to these human-like learning scenarios.

Online learning \cite{Hoi2021OnlineLA} and continual learning \cite{Parisi2018ContinualLL,DeLange2019ACL} are examples of the research fields that address such challenges.
Both of them update a model incrementally by learning from a stream of data, but there are subtle differences in their goals and assumptions.
Most notably, online learning assumes a stationary stream, while continual learning aims to mitigate \emph{catastrophic forgetting} when learning from a non-stationary stream.

Another important body of research is meta-learning  \cite{Hospedales2020MetaLearningIN}.
Unlike traditional machine learning, where a learning algorithm optimizes a model with a training set, meta-learning focuses on optimizing the learning algorithm in a data-driven manner, such that it produces better models than manually crafted learning algorithms.
Therefore, it is often described as ``learning to learn.''
While standard learning involves only a single learning episode, meta-learning consists of multiple learning episodes, which are split into meta-training and meta-test sets.
By utilizing the meta-knowledge extracted from the multiple episodes in the meta-training set, meta-learning aims to improve learning in each episode of the meta-test set.
Meta-learning is also closely related to various other domains, such as few-shot learning \cite{Snell2017PrototypicalNF, Sung2017LearningTC, Li2017MetaSGDLT} and transfer learning \cite{Li2020OnlineMF, Li2019FeatureCriticNF}.

While these learning frameworks have undergone extensive research individually, there has been a recent surge of interest in their synergistic combination, leading to the emergence of novel problem settings and methodologies.
For example, with the help of meta-learning, we can optimize a continual learning algorithm in a data-driven manner, instead of manually designing it.
In this case of \emph{meta-continual learning}, each episode of traditional meta-learning is replaced with a continual learning episode.
Often referred to as ``learning to continually learn,'' meta-continual learning can be found in the evolution of human intelligence.
We could develop our strong continual learning ability over millions of years of evolution since it is crucial for survival and reproduction.
Similarly, if there are multiple continual learning episodes available, we can improve continual learning algorithms by meta-learning.

Another example is \emph{online meta-learning}.
Humans perform extensive meta-learning throughout their lifetime; the more learning experiences we accumulate, the better we become at learning new knowledge and skills.
However, we encounter each learning episode sequentially, not all at once as in traditional meta-learning.
Online meta-learning aims to mimic this sequential meta-learning process by presenting learning episodes in a streaming fashion.
This framework is suitable for applications where the learning episodes are consistently collected, and the learning algorithm should be adapted to the new episodes continuously.

Although each of such combinations possesses unique characteristics and significance, they can be easily confused for several reasons.
First, combining multiple frameworks inevitably introduces additional complexity to the problem formulations.
Second, even if the fundamental structures of problem settings are the same, individual papers often introduce some variations, making them harder to classify.
Lastly, a lack of unified terminology and formulations exacerbates the confusion.

In this context, our primary objective is to present a comprehensive overview of the intersections among online learning, continual learning, and meta-learning.
We establish a unified notation to organize existing learning frameworks into a well-defined taxonomy.
Using our notation, we first define the four basic learning frameworks, i.e., offline learning, online learning, continual learning, and meta-learning.
Then, we categorize their combinations into five major branches: meta-online learning, meta-continual learning, online meta-learning, continual meta-learning, and continual bi-level learning.
For each of these combinatorial learning frameworks, we provide a formal definition and survey relevant papers.

While we do explain the concepts of the basic learning frameworks, this work primarily focuses on their combinations, which currently lack a comprehensive survey.
Although not strictly necessary, referencing existing surveys can facilitate a deeper understanding of the topic.
We recommend consulting the surveys \cite{Hoi2021OnlineLA} for online learning, \cite{Parisi2018ContinualLL, DeLange2019ACL} for continual learning, and \cite{Hospedales2020MetaLearningIN, Vettoruzzo2023AdvancesAC, Vanschoren2019AML} for meta-learning.

In summary, our contribution can be outlined as follows.
First, we define a clear taxonomy for learning frameworks that are combinations of online, continual, and meta-learning.
Second, we provide a comprehensive survey of each category, identifying various research branches.
Third, we explore the remaining challenges and propose potential avenues for future work in this promising field.

The rest of this paper is organized as follows.
In \S\ref{sec:taxonomy}, we present the definitions and taxonomy of the learning frameworks.
We supplement the descriptions with formal algorithms and graphical illustrations of the data structure.
In \S\ref{sec:prelim}-\S\ref{sec:cbl}, we provide a comprehensive survey and discussions of each learning framework.
In \S\ref{sec:app}, we introduce practical applications of the learning frameworks in various domains.
In \S\ref{sec:future}, we address remaining challenges and propose potential research directions.
Lastly, in \S\ref{sec:conclusion}, we conclude with general discussions and future remarks.

\section{Taxonomy of Learning Frameworks}
\label{sec:taxonomy}

\begin{figure*}
  \begin{minipage}[t]{0.32\textwidth}
    \vspace{1pt}
    \begin{subfigure}{\textwidth}
      \centering
      \includegraphics[scale=1.0]{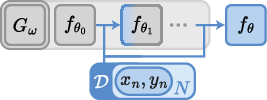}
      \caption*{(a) Offline learning}
    \end{subfigure}

    \vspace{8pt}
    \begin{subfigure}{\textwidth}
      \centering
      \includegraphics[scale=1.0]{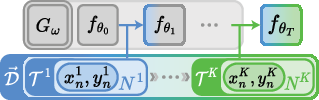}
      \caption*{(c) Offline Continual Learning}
    \end{subfigure}

    \vspace{8pt}
    \begin{subfigure}{\textwidth}
      \centering
      \includegraphics[scale=1.0]{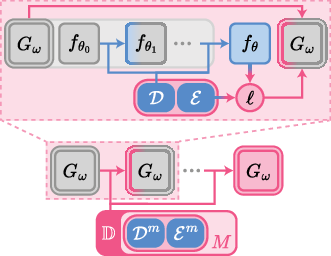}
      \caption*{(f) Meta-Learning}
    \end{subfigure}
  \end{minipage}
  \hfill
  \begin{minipage}[t]{0.32\textwidth}
    \vspace{1pt}
    \begin{subfigure}{\textwidth}
      \centering
      \includegraphics[scale=1.0]{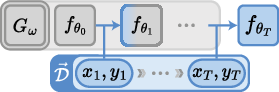}
      \caption*{(b) Online Learning}
    \end{subfigure}

    \vspace{8pt}
    \begin{subfigure}{\textwidth}
      \centering
      \includegraphics[scale=1.0]{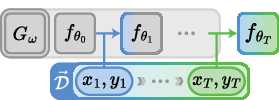}
      \caption*{(d) Online Continual Learning}
    \end{subfigure}

    \vspace{8pt}
    \begin{subfigure}{\textwidth}
      \centering
      \includegraphics[scale=1.0]{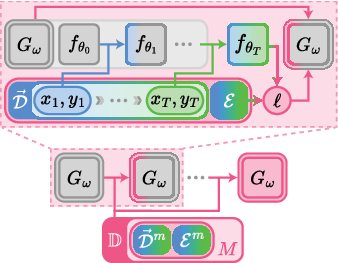}
      \caption*{(g) Meta-Continual Learning}
    \end{subfigure}
  \end{minipage}
  \hfill
  \begin{minipage}[t]{0.32\textwidth}
    \vspace{44pt}
    \begin{subfigure}{\textwidth}
      \centering
      \includegraphics[scale=1.0]{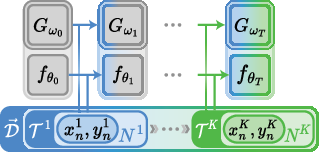}
      \caption*{(e) Continual Bi-Level Learning}
    \end{subfigure}

    \vspace{8pt}
    \begin{subfigure}{\textwidth}
      \centering
      \includegraphics[scale=1.0]{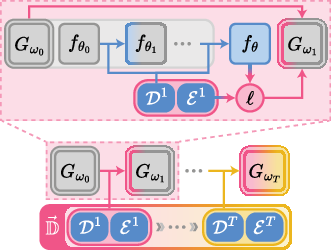}
      \caption*{(h) Continual Meta-Learning}
    \end{subfigure}
  \end{minipage}
\caption{Taxonomy of learning frameworks.}
\label{fig:taxonomy}
\end{figure*}

This section defines the problem formulations of relevant learning frameworks.
We categorize learning frameworks into eight major branches depending on their problem settings: offline learning, online learning, continual learning (CL), meta-learning, meta-online learning (MOL), meta-continual learning (MCL), online meta-learning (OML), and continual meta-learning (CML).
The methodologies to tackle these problems will be described in \S\ref{sec:prelim}-\ref{sec:cbl}.
Fig.~\ref{fig:taxonomy} visualizes the key characteristics and the data structure of each learning framework, and Table \ref{tab:notations} summarizes our unified notations used throughout this paper.

\begin{table}[h]
  \centering
  \caption{Notations}
  \label{tab:notations}
  \begin{tabular}{ll}
    \toprule
    Notation & Meaning \\
    \midrule
    $x$ & An input data \\
    $y$ & A target data \\
    $\ex$ & An example \\
    $\trainset = \trainsetdef$ & A training set \\
    $\trainstream = \trainstreamdef$ & A training stream \\
    $\testset = \testsetdef$ & A test set \\
    $\epi$ or $(\trainstream, \testset)$ & An episode \\
    $\mtrainset = \mtrainsetdef$ & A meta-training set \\
    $\mtrainstream = \mtrainstreamdef$ & A meta-training stream \\
    $\mtestset = \mtestsetdef$ & A meta-test set \\
    $\model_\modelparam$ & Model parameterized by $\modelparam$ \\
    $\learner_\learnerparam$ & Learner parameterized by $\learnerparam$ \\
    $\loss = \Ls(\model_\modelparam, \testset)$ & The loss of $\model_\modelparam$ on $\testset$ \\
    \bottomrule
  \end{tabular}
\end{table}

\begin{table*}
  \centering
  \caption{Comparison of learning frameworks.}
  \label{tab:comparison}
  \vspace{-5mm}
  \begin{tcolorbox}[colback=white, boxrule=0mm, left=0mm, right=0mm, top=2mm, bottom=2mm, boxsep=0mm, colframe=white, sharp corners]
    \centering
  \begin{tabular}{llll}
    \toprule
    Framework & Inner loop data & Outer loop data \\
    \midrule
    Offline learning & Set of examples & N/A \\
    Online learning & Stream of examples & N/A \\
    Online continual learning & Non-stationary stream  of examples & N/A \\
    Offline continual learning & Non-stationary stream of tasks & N/A \\
    Continual bi-level learning & Non-stationary stream of tasks & N/A \\
    Meta-learning & Set of examples & Set of episodes \\
    Meta-online learning & Stream of examples & Set of episodes \\
    Meta-continual learning & Non-stationary stream of examples & Set of episodes \\
    Online meta-learning & Set of examples & Stream of episodes \\
    Continual meta-learning & Set of examples & Non-stationary stream of episodes \\
    \bottomrule
  \end{tabular}
  \end{tcolorbox}
\end{table*}

\subsection{Offline Learning}
\label{sec:tax:offline}
We start by defining the most basic learning framework, which we call \emph{offline learning}, in our terminologies.
In previous works, it has also been called batch learning.
Offline learning aims to produce a \emph{model} from a training set $\trainset = \trainsetdef$ where $x$ and $y$ are the input and target, respectively.
We denote the model $\model_\modelparam: \gX \to \gY$, which is a function parameterized by $\modelparam$.

\paragraph*{Formal Description.}
For a consistent and formal comparison with other frameworks, we introduce the concept of \emph{learner} $\learner_\learnerparam$, to abstract the learning process into a single entity.
In offline learning, $\learner_\learnerparam: \gP((\gX \times \gY)) \to (\gX \to \gY)$ is a functional that takes a training set $\trainset$ and outputs a trained model $\model_\modelparam$, as summarized in Alg.~\ref{alg:offline} and Fig.~\ref{fig:taxonomy}a.
The learner contains everything required for training, such as the model architecture, its initialization scheme, an optimization algorithm, etc.
The learner's parameter $\learnerparam$ consists of the hyperparameters such as the learning rate, which are set by human and remains fixed during learning.
In typical deep learning scenarios, the optimization process inside $\learner$ is performed via mini-batch stochastic gradient descent (SGD).
Since the whole training set is generally too large for optimization, a much smaller subset of examples, i.e., a mini-batch, is sampled at each training step.
The examples inside the mini-batch are forwarded through the current model $\model_\modelparam$, producing a loss term $\loss$.
Then, the gradient $\nabla_\modelparam \loss$ is computed with the backpropagation algorithm, which is then fed into the SGD or a more advanced optimization algorithm (e.g., \cite{Kingma2015AdamAM}) to update $\modelparam$.

\begin{algorithm}[h]
  \caption{Offline learning}
  \label{alg:offline}
  \small
  \begin{algorithmic}[1]
    \Require{training set $\trainset = \trainsetdef$, learner $\learner_\learnerparam$}
    \State $\model_\modelparam \gets \learner_\learnerparam(\emptyset)$ \Comment{Initialize model}
    \While{training}
      \State $\model_{\modelparam} \gets \learner_\learnerparam(\model_\modelparam, \trainset)$ \Comment{Update model}
    \EndWhile
    \State \Return $\model_\modelparam$
  \end{algorithmic}
\end{algorithm}

Note that generative models in unsupervised settings can also be represented by replacing either the input or output with a different type.
For example, the models that generate samples \cite{Goodfellow2014GenerativeAN,Kingma2013AutoEncodingVB,Ho2020DenoisingDP} can be represented as $\model_\modelparam: \{ \emptyset \} \to \gY$, while models that estimate $p(x)$ value \cite{Kingma2013AutoEncodingVB,Du2019ImplicitGA} can be represented as $\model_\modelparam: \gX \to \sR$.

\paragraph*{Evaluation Scheme.}
After learning the training set, the model's performance is evaluated on a test set $\testset = \testsetdef$.
Both $\trainset$ and $\testset$ are typically drawn from the same underlying distribution, but they do not overlap in terms of specific data points.
The most common approach for creating these sets is to initially collect a pool of data and then randomly divide it into training and test sets.
Consequently, merely memorizing each training example will not yield a high score.
Instead, the model should generalize the knowledge from the training set to the test set.
Therefore, the model benefits from a diverse training set that adequately covers the entire data distribution.

\subsection{Online Learning}
\label{sec:tax:online}

While offline learning assumes that the whole training data is available throughout the learning process, \emph{online learning} (or sequential learning) assumes the data is provided as a stream, where each example becomes available sequentially and cannot be accessed again.
Therefore, we use the term training \emph{stream} $\trainstream = \trainstreamdef$, instead of training \emph{set} $\trainset$, to refer to the training data.

\paragraph*{Formal Description.}
As summarized in Alg.~\ref{alg:online-continual} and Fig.~\ref{fig:taxonomy}b, online learning is structured as a loop over the training stream.
When an example $\exn{t}$ is provided at time step $t$, the model from the previous time step should be incrementally updated.
Therefore, the learner's definition has changed to $\learner_\learnerparam: ((\gX \to \gY) \times (\gX \times \gY)) \to (\gX \to \gY)$, a functional taking a model and an example as inputs to output a new model.

\paragraph*{Evaluation Schemes.}
The evaluation scheme of online learning can vary depending on the specific problem settings.
The most popular scheme is to utilize the training stream for testing by evaluating $\model_{\modelparam_{t-1}}$ on the training example $\exn{t}$ before learning it.
However, it is also possible to build a separate test set $\testset$ and evaluate the final model $\model_{\modelparam_T}$ on it, similar to the evaluation of offline learning.

\begin{algorithm}[h]
  \caption{Online learning and online continual learning}
  \label{alg:online-continual}
  \small
  \begin{algorithmic}[1]
    \Require{training stream $\trainstream = \trainstreamdef$, learner $\learner_\learnerparam$}
    \State $\model_{\modelparam_0} \gets \learner_\learnerparam(\emptyset)$ \Comment{Initialize model}
    \For{$t$ in $1, \dots, T$}
      \State $\exn{t} \gets \trainstream[t]$
      \State $\model_{\modelparam_t} \gets \learner_\learnerparam(\model_{\modelparam_{t-1}}, \exn{t})$ \Comment{Update model}
    \EndFor
    \State \Return $\model_{\modelparam_T}$
  \end{algorithmic}
\end{algorithm}

\subsection{Continual Learning}
\label{sec:tax:cl}

Continual learning (CL) shares many similarities with online learning, such as streaming data and incremental model updates.
However, there are some crucial differences in their underlying philosophies.
Most notably, CL assumes a non-stationary training stream where the data distribution evolves over time.
Such distribution shifts are typically achieved by structuring the training stream as a series of distinct \emph{tasks}.
For instance, in Split-MNIST \cite{Zenke2017ContinualLT}, which is a rudimentary CL benchmark, the ten classes of MNIST \cite{LeCun1998GradientbasedLA} are divided into five tasks with two classes each, and the tasks are presented sequentially.
Training a neural network naively on such a non-stationary stream using stochastic gradient descent (SGD) consistently overwrites the knowledge of the previous tasks, which is referred to as \emph{catastrophic forgetting}.
The primary goal of CL is to mitigate such forgetting while continually acquiring new knowledge.
In contrast, online learning does not necessarily assume a non-stationary training stream, and the primary goal is to adapt the model to the streaming data in real-time.

\paragraph*{Offline vs. Online CL.}
CL can be further categorized into offline CL and online CL, depending on how the training data is provided.
Offline CL can be thought of as a series of offline learning with different training sets.
In offline CL, the training stream $\trainstream = (\taskset^k)_{k=1}^K$ comprises $K$ task sets.
At each stage $k$, the entire task set $\taskset^k$ is accessible for updating the model.
Thus, the offline CL process can be described as the learner $\learner_\learnerparam$ taking the previous stage's model $\model_{\modelparam_{k-1}}$ and the current task set $\taskset^k$ to produce an updated model $\model_{\modelparam_{k}}$, as summarized in Alg.~\ref{alg:offline-continual} and Fig.~\ref{fig:taxonomy}c.
On the other hand, in online CL, each example becomes available sequentially, and its overall learning scheme is identical to online learning in Alg.~\ref{alg:online-continual} and Fig.~\ref{fig:taxonomy}d.
The only difference from online learning is the existence of distribution shifts.
Due to the fully streaming nature, online CL is considered more challenging compared to offline CL.
Moreover, online CL can cover a wider range of problem settings, such as task-agnostic CL.

\begin{algorithm}[h]
  \caption{Offline continual learning}
  \label{alg:offline-continual}
  \small
  \begin{algorithmic}[1]
    \Require{training stream $\trainstream = (\taskset^k)_{k=1}^K$ where each task set $\taskset^k = \{(x_n^k, y_n^k)\}_{n=1}^{N_k}$, learner $\learner_\learnerparam$}
    \State $\model_{\modelparam_0} \gets \learner_\learnerparam(\emptyset)$ \Comment{Initialize model}
    \For{$k$ in $1, \dots, K$}
      \State $\taskset^k \gets \trainstream[k]$
      \State $\model_{\modelparam_k} \gets \learner_\learnerparam(\model_{\modelparam_{k-1}}, \taskset^k)$ \Comment{Update model}
    \EndFor
    \State \Return $\model_{\modelparam_K}$
  \end{algorithmic}
\end{algorithm}

\paragraph*{Task-Aware vs. Task-Agnostic CL.}
Depending on the availability of explicit task information, such as the task identity of each example, we can divide CL into task-aware and task-agnostic CL.
Note that offline CL is inherently a task-aware CL setting since the training stream is provided as a sequence of task sets.
In task-aware CL, we can leverage additional task information in various ways.
For example, it can be used for performing special operations at task transition boundaries \cite{Kirkpatrick2016OvercomingCF, Schwarz2018ProgressC} or assigning examples to task-specific components \cite{Rusu2016ProgressiveNN,Aljundi2016ExpertGL}.
In contrast, task-agnostic CL \cite{Aljundi2018TaskFreeCL, Lee2020AND} lacks this task information, rendering many task-aware CL methods impractical or obsolete.
However, task-agnostic CL can represent more diverse problem settings, including scenarios where clear task boundaries do not exist.

\paragraph*{Evaluation Schemes.}
The primary evaluation metric of CL is the performance of the final model ($\model_{\modelparam_K}$ or $\model_{\modelparam_T}$) on a test set $\testset$, which consists of test examples from all tasks.
Therefore, the model needs to retain knowledge from earlier tasks to achieve a high score.
It is also a common practice to measure performance multiple times during training and evaluate various auxiliary metrics, such as forgetting, backward transfer, and forward transfer \cite{LopezPaz2017GradientEM,Rodrguez2018DontFT}.
Although these can be useful for analyzing and comparing the properties of different CL approaches, they should be carefully used in conjunction with the final score since they measure the relative performance change within a CL episode.
For example, the forgetting metric measures the amount of performance degradation of a task after it is learned, but lower forgetting score does not necessarily implies better CL performance.
Specifically, if the initial accuracy of the task is zero, it cannot get any worse, so the forgetting is also zero.

\subsection{Meta-Learning}
\label{sec:tax:meta}

In offline learning, a predefined learning algorithm distills a training set $\trainset$ into a model $\model_\modelparam$.
However, high-dimensional problems inherently require a substantial number of examples to meaningfully specify the relationship between the input and output.
Even with a sufficient number of examples, the optimization process can be slow, necessitating a large number of learning iterations over the training set to find a competent model.
Meta-learning is a data-driven approach to such challenges of offline learning.
By employing a \emph{meta-training set} that comprises multiple learning episodes, meta-learning aims to extract meta-knowledge from it to optimize the learning algorithm, which we denote as learner $\learner_\learnerparam$, to reduce the required number of examples and accelerate adaptation.
The final product of meta-learning is thus not a model but a learner, which is expected to produce competitive models in novel learning episodes.
Therefore, meta-learning is often described as \emph{learning to learn}.
To transfer the meta-learned knowledge to the meta-test episodes, the meta-test episodes cannot be fundamentally different from the meta-training episodes \cite{PAC}.
Therefore, most works assume that both the meta-training and meta-test sets are sampled from the same distribution, similar to how training and test sets in offline learning are sampled from the same distribution.
However, one may relax this assumption and allow the meta-test episodes to be sampled from a different distribution, as long as the meta-training and meta-test sets share key characteristics.

\paragraph*{Formal Description.}
A learning episode $\epi$ is a pair of a training set $\trainset$ and a test set $\testset$.
A meta-training set $\mtrainset = \mtrainsetdef$ consists of $M$ episodes.
Note that, in the meta-learning literature, episodes are more commonly referred to as \emph{tasks}, and the training and test sets are called \emph{support and query sets}, respectively.
However, we stick to the terms episode, training set, and test set to avoid collision with the CL terminology and maintain consistency throughout the paper.
The meta-learning framework is structured as a bi-level optimization problem as summarized in Alg.~\ref{alg:meta} and Fig.~\ref{fig:taxonomy}f.
In the outer loop, a learner is trained on a meta-training set, and in the inner loop, a model is produced from a training set by the learner.
Each inner loop is equivalent to a typical offline learning scenario.
Note that, in Alg.~\ref{alg:meta}, the inner \emph{loop} may not be apparent since we define the learner $\learner_\learnerparam$ to include the whole optimization process for producing a model, which typically consists of multiple iterations over the training set.
At each outer-loop iteration, an episode $\epi$ (more generally, a mini-batch of episodes) is sampled from the meta-training set $\mtrainset$ (Alg.~\ref{alg:meta} L\ref{alg:meta:epi-sample}).
The current learner $\learner_\learnerparam$ then produces a model $\model_\modelparam$ from the training set $\trainset$ (L\ref{alg:meta:inner}), as in offline learning.
This model is evaluated on the corresponding test set to produce the loss $\loss$ (L\ref{alg:meta:loss}), which is used to meta-update the learner's parameter $\learnerparam$ (L\ref{alg:meta:meta-update}).

\begin{algorithm}[h]
  \caption{Meta-learning}
  \label{alg:meta}
  \small
  \begin{algorithmic}[1]
    \Require{
    meta-training set $\mtrainset = \mtrainsetdef$,
    meta-learner $\mlearner$}
    \State $\learner_\learnerparam \gets \mlearner(\emptyset)$ \Comment{Initialize learner}
    \While{meta-training} \label{alg:meta:outer} \Comment{Outer loop}
      \State $\epi \gets \textrm{sample from } \mtrainset$ \label{alg:meta:epi-sample}
      \State $\model_{\modelparam} \gets \learner_\learnerparam(\emptyset)$ \Comment{Initialize model}
      \While{training} \Comment{Inner loop}\label{alg:meta:inner}
        \State $\model_{\modelparam} \gets \learner_\learnerparam(\model_\modelparam, \trainset)$ \Comment{Update model}
      \EndWhile
      \State $\loss \gets \Ls(\model_{\modelparam}, \testset)$ \label{alg:meta:loss} \Comment{Evaluate model}
      \State $\learner_\learnerparam \gets \mlearner(\learner_\learnerparam, \loss)$ \label{alg:meta:meta-update} \Comment{Update learner}
    \EndWhile
    \State \Return $\learner_\learnerparam$
  \end{algorithmic}
\end{algorithm}

\paragraph*{Evaluation Scheme.}
Once a meta-trained learner is obtained as a result of meta-training, it is evaluated on the meta-test set, which consists of novel episodes.
For each episode in the meta-test set, the learner produces a trained model from the training set, which is subsequently evaluated on the test set.
The average of these test scores is used as an evaluation metric for meta-learning algorithms.
Since the goal of the meta-test is to evaluate the learning capability of the learner, not the specific knowledge from the meta-training episodes, the meta-training and meta-test sets generally do not have any overlap in the underlying data.
For example, in the Omniglot benchmark \cite{Lake2011OneSL}, 1,623 character classes are first split into two groups with 963 and 660 classes each, which then constitute the meta-training and meta-test episodes, respectively.
Therefore, simply memorizing the meta-training data does not help achieve a high score in the meta-test set.
Instead, the learner must generalize the meta-knowledge extracted from the meta-training episodes to the meta-test episodes.
Just as offline learning benefits from a diverse training set that covers the entire data distribution, meta-learning benefits from a diverse meta-training set.

\subsection{Meta-Online Learning \& Meta-Continual Learning}
\label{sec:tax:mcl-mol}

If meta-learning is the process of optimizing a learning algorithm for offline learning, a logical extension would be to adapt it to optimize learning algorithms for online learning or continual learning.
We refer to these frameworks as \emph{meta-online learning} (MOL) and \emph{meta-continual learning} (MCL), respectively.
The goals of MOL and MCL can be summarized as \emph{learning to sequentially learn} and \emph{learning to continually learn}.

\paragraph*{Motivation.}
Humans, along with many other intelligent animals, exhibit a remarkable level of online and continual learning ability.
Nevertheless, our learning abilities are not flawless; they are rather specialized in specific domains critical for survival and reproduction, such as language, facial recognition, and sensory-motor coordination.
When it comes to other tasks, like memorizing random digits, we often face much trouble learning new knowledge incrementally.
Considering that infants rapidly accumulate knowledge with minimal forgetting, it seems plausible that humans have likely developed their capacity for continual and online learning through the process of biological evolution rather than solely within an individual's lifetime.
In this perspective, each lifetime can be viewed as a CL episode, and the evolutionary process can be regarded as a form of meta-continual learning or meta-online learning.

\paragraph*{Formal Description.}
Since episodes in MCL are generally online CL episodes, MCL can be summarized as Alg.~\ref{alg:meta-continual} and Fig.~\ref{fig:taxonomy}g, where the inner loop of meta-learning (Alg.~\ref{alg:meta}) is replaced by incremental updates over a training stream $\trainstream$.
The trained model is then evaluated on a test set to produce a loss $\loss$, and the learner $\learner_\learnerparam$ is updated to reduce the loss, which is often done by gradient descent.
In the case of MOL, there is no separate test set in each episode, as the training stream also serves as test data in typical online learning.
In each inner loop iteration of MOL, the model $\model_{\modelparam_{t-1}}$ from the previous iteration is evaluated on the current example $\exn{t}$, yielding the loss $\loss_t$.
After training, these losses are aggregated and fed into the meta-update rule to update the learner $\learner_\learnerparam$.
Note that this description is one possible form of meta-training processes.
As mentioned in \S\ref{sec:tax:meta}, the meta-learning framework allows variations in the meta-training scheme, which also applies to MCL and MOL settings.

\begin{algorithm}[h]
  \caption{Meta-online learning}
  \label{alg:meta-online}
  \small
  \begin{algorithmic}[1]
    \Require{
    meta-training set $\mtrainset = \{\trainstream^m\}_{m=1}^M$, 
    meta-learner $\mlearner$}
    \State $\learner_\learnerparam \gets \mlearner(\emptyset)$ \Comment{Initialize learner}
    \While{meta-training} \Comment{Outer loop}
      \State $(\trainstream, \testset) \gets \textrm{sample from } \mtrainset$ \label{alg:meta-online:epi-sample}
      \State $\model_{\modelparam_0} \gets \learner_\learnerparam(\emptyset)$ \Comment{Initialize model}
      \For{$t$ in $1, \dots, |\trainstream|$} \Comment{Inner loop}
        \State $\exn{t} \gets \trainstream[t]$
        \State $\loss_t \gets \Ls(\model_{\modelparam_{t-1}}, \exn{t})$ \Comment{Evaluate model}
        \State $\model_{\modelparam_t} \gets \learner_\learnerparam(\model_{\modelparam_{t-1}}, \exn{t})$ \Comment{Update model}
      \EndFor
      \State $\learner_\learnerparam \gets \mlearner(\learner_\learnerparam, (\loss_t)_{t=1}^{|\trainstream|})$ \label{alg:mcl-prob:meta-update} \Comment{Update learner}
    \EndWhile
    \State \Return $\learner_\learnerparam$
  \end{algorithmic}
\end{algorithm}

\begin{algorithm}[h]
  \caption{Meta-continual learning}
  \label{alg:meta-continual}
  \small
  \begin{algorithmic}[1]
    \Require{
    meta-training set $\mtrainset = \{(\trainstream^m, \testset^m)\}_{m=1}^M$, 
    meta-learner $\mlearner$}
    \State $\learner_\learnerparam \gets \mlearner(\emptyset)$ \Comment{Initialize learner}
    \While{meta-training} \Comment{Outer loop}
      \State $(\trainstream, \testset) \gets \textrm{sample from } \mtrainset$ \label{alg:meta-continual:epi-sample}
      \State $\model_{\modelparam_0} \gets \learner_\learnerparam(\emptyset)$ \Comment{Initialize model}
      \For{$t$ in $1, \dots, |\trainstream|$} \Comment{Inner loop}
        \State $\exn{t} \gets \trainstream[t]$
        \State $\model_{\modelparam_t} \gets \learner_\learnerparam(\model_{\modelparam_{t-1}}, \exn{t})$ \Comment{Update model}
      \EndFor
      \State $\loss \gets \Ls(\model_{\modelparam_{|\trainstream|}}, \testset)$ \label{alg:meta-continual:loss} \Comment{Evaluate model}
      \State $\learner_\learnerparam \gets \mlearner(\learner_\learnerparam, \loss)$ \label{alg:meta-continual:meta-update} \Comment{Update learner}
    \EndWhile
    \State \Return $\learner_\learnerparam$
  \end{algorithmic}
\end{algorithm}

\paragraph*{Evaluation Schemes.}
In both MOL and MCL, the meta-trained learner $\learner_\learnerparam$ is evaluated on a meta-test set $\mtestset$.
For each episode in the meta-test set, the learner iteratively updates a model with the examples from the training stream.
The model is then evaluated on a test set, or in MOL, the model can be simultaneously evaluated on the training stream without employing a test set.
The evaluation results are then aggregated to be used as the final performance metric.

\subsection{Online Meta-Learning \& Continual Meta-Learning}
\label{sec:tax:oml-cml}

While MCL and MOL replace the inner loop of meta-learning with a stream of examples, online meta-learning (OML) and continual meta-learning (CML) replace the \emph{outer loop} of meta-learning with a stream of \emph{episodes}.
Typically, these episodes are offline learning episodes, each of which comprises a training set and a test set.
The subject of incremental updates is a learner.
Therefore, OML and CML can be described as \emph{sequentially learning to learn} and \emph{continually learning to learn}, respectively.
Although we mostly consider offline learning episodes in the following description, other types of learning episodes can be employed, such as online learning or reinforcement learning.

\paragraph*{Motivation.}
OML settings are often inspired by humans' meta-learning capability being continuously improved throughout their lifetime.
For example, learning a mathematical topic can be challenging at first, but as we learn many more topics, we become more adept at learning new ones.
On the other hand, the focus of CML is to prevent forgetting how to learn.
As an example, consider an agent operating in an evolving environment where the dynamics consistently change.
The agent should learn new optimal behaviors when the dynamics change.
However, if new dynamics have been encountered before, it should be able to adapt much more quickly, compared to learning from scratch.
Therefore, the CML setting is sometimes referred to as \emph{faster remembering} \cite{He2019TaskAC}.

\paragraph*{Formal Description.}
Both OML and CML are summarized as Alg.~\ref{alg:continual-meta} and Fig.~\ref{fig:taxonomy}h, whose overall structure largely resembles that of meta-learning (Alg.~\ref{alg:meta}).
The key difference is that the meta-training set $\mtrainset = \mtrainsetdef$ is replaced with a meta-training stream $\mtrainstream = \mtrainstreamdef$.
As a result, the outer loop now iterates over each episode in the meta-training stream.
Theoretically, one can distinguish between CML and OML settings by examining the stationarity of the meta-training stream.
In CML, the episode distribution shifts during meta-training, while it remains stationary in OML.
However, determining whether the meta-training stream is stationary may not always be straightforward, especially when the number of episodes is small.
In such cases, examining the evaluation scheme can help clarify the problem setting.

\begin{algorithm}[h]
  \caption{Online meta-learning and continual-meta learning}
  \label{alg:continual-meta}
  \small
  \begin{algorithmic}[1]
    \Require{meta-training stream $\mtrainstream = \mtrainstreamdef$, meta-learner $\mlearner$}
    \State $\learner_{\learnerparam_0} \gets \mlearner(\emptyset)$ \Comment{Initialize learner}
    \For{$u$ in $1, \dots, U$} \Comment{Outer loop}
      \State $\model_\modelparam \gets \learner_{\learnerparam_{u-1}}(\emptyset)$ \Comment{Initialize model}
      \While{training} \Comment{Inner loop}
        \State $\model_{\modelparam} \gets \learner_{\learnerparam_{u-1}}(\model_\modelparam, \trainset^u)$ \Comment{Update model}
      \EndWhile
      \State $\loss \gets \Ls(\model_{\modelparam^u}, \testset^u)$ \Comment{Evaluate model}
      \State $\learner_{\learnerparam_u} \gets \mlearner(\learner_{\learnerparam_{u-1}}, \loss)$ \Comment{Update learner}
    \EndFor
    \State \Return $\learner_{\learnerparam_U}$
  \end{algorithmic}
\end{algorithm}

\paragraph*{Evaluation Schemes.}
Since CML and OML are also meta-learning approaches, the subject of evaluation is the learner, not the model.
In CML, the meta-trained learner is evaluated on a meta-test set, whose episodes are similar to those in the meta-training stream.
On the other hand, OML evaluation does not involve a separate meta-test set.
During meta-training, the performance of the learner in each new meta-training episode is used as the evaluation metric.

\subsection{Continual Bi-Level Learning}
\label{sec:tax:cbl}
There is another group of approaches that leverage meta-learning within the CL problem setting.
We coin the term \emph{continual bi-level learning} (CBL) to refer to such approaches.
In terms of the problem setting, CBL is not different from CL.
As in CL, the end goal is to produce a model by sequentially training it on a non-stationary data stream, and there is no concept of meta-training or meta-test.
CBL differs from CL in that the learning algorithm is also updated along with the model as depicted in Fig.~\ref{fig:taxonomy}e, thus the name bi-level learning.
In other words, meta-learning is concurrently performed during CL, to improve the CL performance dynamically.
This contrasts with the standard CL, where the learning algorithm stays the same, regardless of the amount of learned knowledge.

\paragraph*{Formal Description.}
In most cases, CBL follows an offline, task-aware CL setting, where the training stream $\trainstream$ is a sequence of task sets.
Therefore, the overall structure of CBL in Alg.~\ref{alg:continual-bi} closely resembles offline CL, which is summarized in Alg.~\ref{alg:offline-continual}.
The key difference can be found in L\ref{alg:continual-bi:update}, where the learner is also updated in addition to the model.
As a result, the learning dynamics change as the training progresses.
Although the learner is updated during training, only the model is treated as the final product (L\ref{alg:continual-bi:return}).

\begin{algorithm}[h]
  \caption{Continual bi-level learning}
  \label{alg:continual-bi}
  \small
  \begin{algorithmic}[1]
    \Require{training stream $\trainstream = (\taskset^k)_{k=1}^K$ where each task set $\taskset^k = \{(x_n^k, y_n^k)\}_{n=1}^{N_k}$, initial model $\model_{\modelparam_0}$, initial learner $\learner_{\learnerparam_0}$}
    \For{$k$ in $1, \dots, K$}
      \State $\taskset^k \gets \trainstream[k]$
      \State $(\model_{\modelparam_k}, \learner_{\learnerparam_k}) \gets \learner_{\learnerparam_{k-1}}(\model_{\modelparam_{k-1}}, \taskset^k)$ \label{alg:continual-bi:update} \Comment{Update model and learner}
    \EndFor
    \State \Return $\model_{\modelparam_K}$ \label{alg:continual-bi:return}
  \end{algorithmic}
\end{algorithm}

\paragraph*{Evaluation Schemes.}
The evaluation scheme for CBL basically follows CL.
The trained model is evaluated on a test set containing test examples for the tasks learned during training.

\section{Preliminary}
\label{sec:prelim}

The deep learning community has witnessed a surge in research within the fields of continual learning and meta-learning.
In this section, we offer an overview of the methods proposed, which is essential for understanding the upcoming sections.
We begin by introducing previous works on meta-learning (\S\ref{sec:prelim:ml}), followed by a discussion of continual learning (\S\ref{sec:prelim:cl}).
It is important to note that the review of this section is confined to methods directly relevant to following sections, with other significant contributions omitted.

\subsection{Meta-Learning Methods}
\label{sec:prelim:ml}

Meta-learning methods can be roughly classified into \emph{gradient-based}, \emph{model-based}, and \emph{metric-based} approaches \cite{Lee2018GradientBasedMW, yao2020automated, Hospedales2020MetaLearningIN}.
To conduct a systematic examination, We adhere to this taxonomy and present relevant methods within each respective category.
For readers seeking a comprehensive overview of meta-learning, we direct them to a previous survey \cite{Hospedales2020MetaLearningIN}.

\subsubsection{Gradient-Based Meta-Learning}
\label{sec:prelim:ml:opt}

Gradient descent is a fundamental optimization strategy in deep learning.
Understandably, there has been significant research focused on learning the gradient descent process, known as gradient-based meta-learning \cite{Andrychowicz2016LearningTL, li2017learning, Finn2017ModelAgnosticMF, Antoniou2018HowTT}.
In these approaches, both inner and outer loops consist of the gradient descent processes.
Within the outer loop, the gradient descent algorithm is iteratively improved, while in the inner loop, the algorithm is used to optimize a model.
Therefore, gradient-based meta-learning is also referred to as \emph{bi-level optimization}.
Formally, the learner parameter $\learnerparam$ in these approaches includes optimizable factors associated with the effectiveness of the optimization algorithm, such as model initialization and learning rate.

\paragraph*{MAML \cite{Finn2017ModelAgnosticMF}.}
Model-Agnostic Meta-Learning (MAML) is a well-known gradient-based approach.
In the context of MAML, the extracted meta-knowledge from episodes takes the form of \emph{initialization} of the model parameters, represented as $\learnerparam = \{ \theta_0 \}$.
When faced with an episode, MAML updates the model using gradient descent on the training set.
Following this, the test set is employed to assess the updated parameters, which generates the meta-loss.
By identifying the optimal initialization, MAML aims to rapidly \emph{adapt} the model during the meta-test phase using only a few gradient updates.
Subsequent studies have extended MAML by augmenting the learner parameter $\learnerparam$ with additional components, such as the learning rate of the inner gradient descent \cite{Li2017MetaSGDLT, Antoniou2018HowTT}.

\paragraph*{Challenges of Bi-Level Optimization.}
However, these approaches may face challenges as they require conducting gradient descent throughout the entire adaptation process.
The challenge involves computing second-order derivatives of the loss function, as well as retaining the entire computation graph of the inner loop.
These limitations have been criticized for hindering the scalability of bi-level optimization \cite{Antoniou2018HowTT, Hospedales2020MetaLearningIN}.
To alleviate the limitations, a number of studies have proposed alternative methods to enhance the scalability \cite{Rajeswaran2019MetaLearningWI, Flennerhag2019MetaLearningWW, Shaban2019TruncatedBF, Bertinetto2018MetalearningWD, Shin2021LargeScaleMW}.

\paragraph*{Approximating Second-Order Derivatives.}
The original MAML paper \cite{Finn2017ModelAgnosticMF} addressed approximating second-order derivatives with first-order derivatives.
Additionally, several studies have introduced alternative methods for approximating the derivatives \cite{Nichol2018OnFM, Rajeswaran2019MetaLearningWI}.
For instance, the Reptile algorithm \cite{Nichol2018OnFM} employs a strategy where, after an adaptation process within an episode resulting in $\theta_L$, the initial model parameter $\theta_0$ is adjusted slightly towards the adapted parameter $\theta_L$.
This algorithm eliminates the necessity for meta-loss computation, rendering the division of training and test sets unnecessary.

\paragraph*{Reducing Adapting Parameters.}
On the other hand, there has been a line of research aiming to reduce the proportion of the model parameters that undergo adaptation \cite{Zintgraf2018FastCA, Javed2019MetaLearningRF, Raghu2019RapidLO}.
These methods typically separate the parameters into two sets: one set is updated only in the outer loop, while the other set is updated in both the inner and outer loops.
By reducing the computation required for calculating second-order derivatives, this approach ha gained wide usage in the studies we will discuss in this paper.
We delve further into the method of \cite{Javed2019MetaLearningRF} in \S\ref{sec:mcl-mol}.

\subsubsection{Model-Based Meta-Learning}
\label{sec:prelim:ml:model}

RNNs inherently adapt over time, processing a sequence of data.
Likewise, model-based meta-learning approaches leverage the forward pass of neural networks to model the learning process \cite{Ravi2016OptimizationAA, Hochreiter2001LearningTL, Mishra2017ASN, Santoro2016MetaLearningWM, Munkhdalai2017MetaN}.
During the inner loop, a training set is provided as input to a neural network, generating embedded states.
These embedded states are subsequently processed to create a meta-loss with a test set.
In the outer loop, the neural network is improved by backpropagating the meta-loss.
In a formal description, the learner parameter $\learnerparam$ in model-based approaches includes the parameters of the neural network.

\subsubsection{Metric-Based Meta-Learning}
\label{sec:prelim:ml:metric}
Rooted in traditional similarity-based machine learning methods like k-nearest neighbors (k-NN) and kernel methods, metric-based approaches learn embeddings and metrics for comparing data points.
These approaches utilize neural networks as metric functions to compare the embeddings of examples in the training and test sets, enabling predictions to be made regarding the test set \cite{Koch2015SiameseNN, NIPS2016_90e13578, Snell2017PrototypicalNF, Sung2017LearningTC, Satorras2017FewShotLW}.

\paragraph*{Prototypical Network \cite{Snell2017PrototypicalNF}.}
Prototypical Network (PN) employs a neural network that maps each example into an embedding space.
When PN encounters the training set, it utilizes the network to derive embeddings of training examples.
These embeddings are then averaged on a class-wise basis to obtain representative values for the classes, referred to as \emph{prototypes}.
To predict the class membership of test examples, PN measures the distances between the embeddings of each test example and the prototypes of all classes.

\subsection{Continual Learning Methods}
\label{sec:prelim:cl}

CL methods can be broadly classified into three categories: regularization, replay, and dynamic architecture \cite{Parisi2018ContinualLL, DeLange2019ACL}.
Following this taxonomy, we review each category and introduce approaches that are relevant to this paper.
For more in-depth review on CL, we recommend referring to \cite{Parisi2018ContinualLL, DeLange2019ACL}.

\subsubsection{Regularization-Based Continual Learning}
\label{sec:prelim:cl:reg}

One potential way to prevent catastrophic forgetting is to adjust the gradient in directions that do not impair performance on previous tasks.
In this context, regularization-based approaches aim to counteract catastrophic forgetting by introducing supplementary loss terms into the objective function \cite{Kirkpatrick2016OvercomingCF, Zenke2017ContinualLT, Schwarz2018ProgressC}.
These additional loss terms interfere with the model update process, thereby safeguarding previously acquired knowledge.
One such example is the Elastic Weight Consolidation (EWC) \cite{Kirkpatrick2016OvercomingCF, Schwarz2018ProgressC}.
EWC identifies critical model parameters that are essential for preserving knowledge related to past tasks, preventing substantial changes to them.
This method recognizes parameters that have experienced significant updates during the learning of previous tasks as pivotal for maintaining performance on earlier tasks.
EWC employs the Fisher information matrix as a weighting factor to quantify the magnitude of updates for each parameter.
A noteworthy advantage of regularization-based approaches lies in their versatility, as they can be seamlessly integrated with other CL methods by simply appending an extra loss term.

\subsubsection{Replay-Based Continual Learning}
\label{sec:prelim:cl:replay}

Another way to circumvent catastrophic forgetting is to store and replay past examples during training, known as replay-based continual learning.
These approaches make use of a memory buffer to store past examples \cite{Rebuffi2016iCaRL, LopezPaz2017GradientEM, Rolnick2018ER, Riemer2018LearningTL, Chaudhry2018EfficientLL, Aljundi2019GradientBS} or a generative model \cite{Shin2017ContinualLW, Lesort2018GenerativeMF, Sun2019LAMOL} to synthesize such examples.

\paragraph*{GEM \cite{LopezPaz2017GradientEM}.}
Gradient Episodic Memory (GEM) preserves a subset of examples from all previously encountered tasks in a replay buffer.
At every step, the model is evaluated on these examples to approximate the gradients of the current parameter w.r.t. each previous task.
The gradient w.r.t. new training data is modified to yield non-negative dot-products with the gradients w.r.t. previous tasks.
This technique prevents later gradient descents increasing the losses of earlier tasks.

\paragraph*{Generative Replay \cite{Shin2017ContinualLW}.}
Generative replay presents an alternative method wherein an additional generative model is used to synthesize past examples, instead of storing them directly in the replay buffer \cite{Shin2017ContinualLW, Wu2018IncrementalCL, vandeVen2018GenerativeRW, Sun2019LAMOL}.
In their paper \cite{Shin2017ContinualLW}, the authors utilized a \emph{generator} based on Generative Adversarial Nets (GAN) \cite{Goodfellow2014GenerativeAN} in conjunction with a \emph{solver}.
When faced with a new task during training, synthetic past examples are generated from a generator trained on past tasks.
Subsequently, synthetic labels are assigned to the examples using the solver trained on past tasks.
These synthetic examples are then merged with examples of current task, creating a combined dataset for training a new generator and solver.
Generative replay aims to mitigate the need for continually increasing memory usage in previous replay-based approaches.

\subsubsection{Dynamic Architecture Continual Learning}
Various approaches have explored CL capability of dynamic architectures \cite{Rusu2016ProgressiveNN, Aljundi2016ExpertGL, Yoon2017LifelongLW, Lee2020AND}, which assign task-specific network parameters to learn new tasks, rather than adhering to static architectures.
However, these approaches have received relatively less attention within the scope of this paper.
Consequently, further investigation and review of this category are excluded at this stage.

\begin{figure*}
  \centering
  \begin{tikzpicture}
      % Define colors
      \definecolor{categorycolor}{RGB}{241, 241, 242}
      \definecolor{subcatcolor}{RGB}{241, 241, 242}

      Shadow properties
      \tikzset{
        shadowedbox/.style={
            preaction={
                fill=black,
                opacity=0.15,
                transform canvas={
                    shift={(0.5mm,-0.5mm)}
                }
            },
            rounded corners=5pt,
            fill=subcatcolor,
            draw=none,
        }
      }

      % Draw main boxes for categories without borders and with shadows
      \draw[shadowedbox] (0,0) rectangle (5.3,8);
      \node at (2.65, 7.4) {\begin{minipage}{4.5cm}\centering \textbf{Meta-online learning \\ \& Meta-continual learning}\end{minipage}};
      \draw[black] (0.2, 6.9) -- (5.1, 6.9);

      \draw[shadowedbox] (5.8,0) rectangle (11.1,8);
      \node at (8.45, 7.4) {\begin{minipage}{4.5cm}\centering \textbf{Online meta-learning \\ \& Continual meta-learning}\end{minipage}};
      \draw[black] (6.0, 6.9) -- (10.9, 6.9);

      \draw[shadowedbox] (11.6,0) rectangle (16.9,8);
      \node at (14.25, 7.4) {\begin{minipage}{4.5cm}\centering \textbf{Continual bi-level learning}\end{minipage}};
      \draw[black] (11.8, 7.1) -- (16.7, 7.1);

      % Draw sub-boxes for Category 1 with shadows
      \draw[shadowedbox] (0.2,5.2) rectangle (5.1,6.8);
      \node at (2.65, 6.4) {\textbf{Stochastic gradient descent}};
      \node at (2.65, 5.8) {\cite{Javed2019MetaLearningRF, Beaulieu2020LearningTC, AlShedivat2017ContinuousAV, Kim2020APG}};

      \draw[shadowedbox] (0.2,3.1) rectangle (5.1,4.7);
      \node at (2.65, 4.3) {\textbf{Sequential Bayesian update}};
      \node at (2.65, 3.7) {\cite{Snell2017PrototypicalNF, Banayeeanzade2021GenerativeVD, Harrison2018MetaLearningPF, Lee2024LearningTC}};

      \draw[shadowedbox] (0.2,0.6) rectangle (5.1,2.6);
      \node at (2.65, 2.2) {\textbf{Sequence modeling}};
      \node at (2.65, 1.4) {\begin{minipage}{4.5cm}\centering \cite{lee2023recasting, Ravi2016OptimizationAA, Santoro2016MetaLearningWM, Mishra2017ASN, Duan2016RL2FR, Nagabandi2018LearningTA, Lu2023StructuredSS} \end{minipage}};

      % Draw sub-boxes for Category 2 with shadows
      \draw[shadowedbox] (6.0,5.2) rectangle (10.9,6.8);
      \node at (8.45, 6.4) {\textbf{Unitary Initialization}};
      \node at (8.45, 5.8) {\cite{Finn2019OnlineM, Acar2021MemoryEO, Yap2020AddressingCF, He2019TaskAC, Caccia2020OnlineFA, Clark2022MetaLearningFW}};

      \draw[shadowedbox] (6.0,3.1) rectangle (10.9,4.7);
      \node at (8.45, 4.3) {\textbf{Mixture of Initialization}};
      \node at (8.45, 3.7) {\cite{Jerfel2018ReconcilingMA, Nagabandi2018DeepOL, Zhang2021VariationalCB}};

      \draw[shadowedbox] (6.0,1.0) rectangle (10.9,2.6);
      \node at (8.45, 2.2) {\textbf{Compositional Initialization}};
      \node at (8.45, 1.6) {\cite{Yao2020OnlineSM, pmlr-v202-wu23d}};

      % Draw sub-boxes for Category 3 with shadows
      \draw[shadowedbox] (11.8,4.9) rectangle (16.7,6.9);
      \node at (14.25, 6.5) {\textbf{Bi-level optimization}};
      \node at (14.25, 5.7) {\begin{minipage}{4.5cm}\centering \cite{Riemer2018LearningTL, Gupta2020LaMAMLLM, Rajasegaran2020iTAMLAI, Volpi2020ContinualAO, Joseph2020IncrementalOD, Obamuyide2019MetaLearningIL, Wang2020EfficientML, Wu2021CurriculumMetaLF, Jin2021LearnCG, wu2024meta} \end{minipage}};

      \draw[shadowedbox] (11.8,3.1) rectangle (16.7,4.7);
      \node at (14.25, 4.3) {\textbf{Hypernetwork}};
      \node at (14.25, 3.7) {\cite{Oswald2019ContinualLW, Hu2018OvercomingCF, Joseph2020MetaConsolidationFC, Ehret2020ContinualLI, Chandra2022ContinualLW, Hemati2023PartialHF}};

      \draw[shadowedbox] (11.8,0.2) rectangle (16.7,2.9);
      \node at (14.25, 2.5) {\textbf{Pseudo-meta-training set}};
      \node at (14.25, 1.4) {\begin{minipage}{4.5cm}\centering \cite{Tao2020FewShotCL, Cheraghian2021SemanticawareKD, Dong2021FewShotCL, Zhou2022ForwardCF, Cheraghian2021SynthesizedFB, Mazumder2021FewShotLL, Hersche2022ConstrainedFC, Zhu2021SelfPromotedPR, Zhang2021FewShotIL, Zhu2023FewshotIL, Zhao2023FewShotCL, Zhuang2023GKEALGK, Chi2022MetaFSCILAM, Zhou2022FewShotCL, PrezRa2020IncrementalFO, Li2020IncrementalFO, Dong2022IncrementalDETRIF, Cheng2022MetaLearningBasedIF, Yin2022SylphAH, Choi2023IncrementalFO} \end{minipage}};
      
  \end{tikzpicture}
  \caption{Prior studies for each category}
  \label{fig:reference_indices}
\end{figure*}
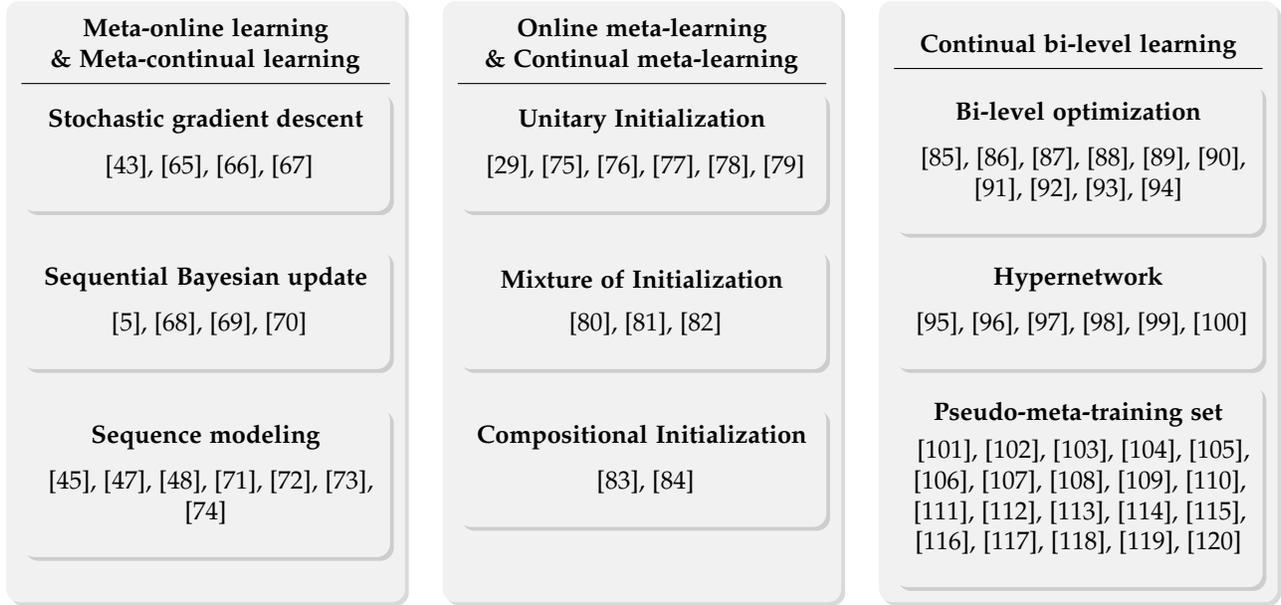

\section{Meta-Online Learning \& Meta-Continual Learning}
\label{sec:mcl-mol}
In this section, we categorize meta-online learning (MOL) and meta-continual learning (MCL) approaches based on their methodological characteristics.
Although MOL and MCL are distinct problem settings, we find methods proposed for one setting are often applicable to the other.
Moreover, several meta-learning approaches, especially in the few-shot learning literature, are also compatible with MOL or MCL settings.
Therefore, we refrain from drawing strict boundaries between the problem settings and focus on what is considered \emph{learning} in each method.
We classify existing approaches into three distinct categories, depending on the primary mechanism to update the model in the inner loop: stochastic gradient descent, sequential Bayesian update, and sequence modeling.

\subsection{Learning as Stochastic Gradient Descent}
\label{sec:mcl-mol:sgd}

Stochastic gradient descent (SGD) has served as the workhorse of deep learning in offline settings and has consequently been embraced as the primary learning mechanism in various other learning scenarios, including MOL, MCL, and meta-learning.
In meta-learning, this direction is often referred to as the gradient-based approach.
MAML \cite{Finn2017ModelAgnosticMF} is a representative example, with the key idea of meta-learning the initial parameters of a model.
The meta-learned initial parameters are regarded as a good starting point that can facilitate the SGD-based optimization process.

Following a similar logic, MAML can be adopted as an MOL or MCL method, i.e., meta-learning a good initialization that can facilitate online or continual learning.
However, since naively applying MAML exposes all parameters to SGD updates, it may result in excessive plasticity, failing to prevent forgetting.

Online aware meta-learning \cite{Javed2019MetaLearningRF} is an MCL approach that can be a solution to this problem.
In the inner loop of this approach, only a couple of topmost layers are updated and the rest is treated as a fixed encoder.
The encoder and the initial parameters of the topmost layers are updated in the outer loop.

In this line of research, ANML \cite{Beaulieu2020LearningTC} is another attempt to balance stability and plasticity in MAML-based approaches.
Instead of freezing the lower encoder part of the model as in \cite{Javed2019MetaLearningRF}, it introduces a separate network called the neuromodulatory network, which remains frozen during the inner loop.
The output of this network has the same shape as the output of the encoder and is passed through the sigmoid function to limit the values in 
$[0, 1]$.
It is then multiplied to the output of the encoder, selectively gating some features.

\begin{figure}
  \includegraphics[width=\linewidth]{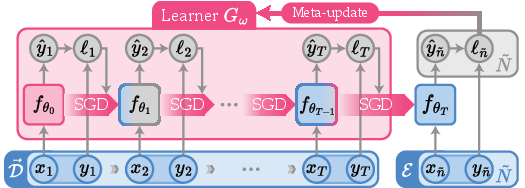}
  \caption{
    Learning as stochastic gradient descent (MCL setting).
  }
  \label{fig:mcl:sgd}
\end{figure}

The core logic of \cite{Javed2019MetaLearningRF} and \cite{Beaulieu2020LearningTC} is illustrated in Fig.~\ref{fig:mcl:sgd}.
Note that the model parameter $\modelparam_t$ only refers to the \emph{fast weights} which are updated inside the inner loop.
The entire model $\model_{\modelparam_t}$ can contain other meta-learned components, or \emph{slow weights}, which are part of the learner's parameter $\learnerparam$.
In the inner loop, $\theta_t$ is sequentially updated via SGD.
In MCL settings, the trained model is evaluated on a test set $\testset$, yielding the meta-loss $\loss_{\tilde n}$ for each test example $\tilde n$.
The learner's parameter $\learnerparam$, which typically comprises the initial model parameter $\theta_0$ and other meta-learned components, is meta-updated to reduce the meta-loss.
In MOL settings, the training loss $\loss_{1:T}$ also serves as the meta-loss, without introducing a separate test set.

A strength of these approaches is that they are often compatible with many existing CL approaches that also rely on SGD as their primary learning mechanism.
However, there are some downsides coming from their roots in MAML.
For instance, computing the gradients of the initial parameters requires computing the second-order gradient through the entire inner loop.
This can entail many practical issues, including heavy computational costs and vanishing/exploding gradients.
For this reason, \cite{Javed2019MetaLearningRF} and \cite{Beaulieu2020LearningTC} approximate the second-order gradient by segmenting the inner loop into smaller chunks, e.g., five iterations, and computing the truncated second-order gradient for each of them.
In addition, manipulating the initialization point alone may not provide sufficient control over the learning dynamics of SGD in continual learning scenarios.\footnote{
  The universal approximation theorem of MAML \cite{Finn2018Universality} does not apply to CL episodes.
}
Since SGD can be considered a greedy algorithm that follows the steepest direction at the current parameter w.r.t. given data, it can still cause catastrophic forgetting when applied on a non-stationary training stream.

\subsection{Learning as Sequential Bayesian Update}
\label{sec:mcl-mol:bayes}

Unlike SGD, which originated from offline settings, the Bayesian framework offers a solid theoretical basis for sequential updates.
From the Bayesian perspective, \emph{learning} implies using the Bayes rule to combine information from data with the prior belief to produce the posterior belief, i.e., posterior $\propto$ prior $\times$ likelihood.
In sequential setups, the posterior belief from the previous time step becomes the prior belief at the current time step.
Therefore, we can formulate online or continual learning as the sequential Bayesian updates of the posterior of some latent variable.

Many CL approaches, especially the regularization-based ones, share this Bayesian perspective \cite{Kirkpatrick2016OvercomingCF,Zenke2017ContinualLT,Chaudhry2018RiemannianWF,Nguyen2017VariationalCL,Farquahr2019Bayesian}.
However, they consider the posterior of neural network parameters, which is generally intractable and requires extensive approximations.
Consequently, they end up with SGD-based update rules accompanied by some regularization tricks, which are often far from ideal Bayesian updates.

On the other hand, once we look away from deep learning, many statistical models are capable of efficient computation of sequential Bayesian updates.
Especially, models that naturally come with an exponential family posterior (e.g., Gaussian) are particularly valuable.
Exponential family distributions are the only class of distributions that always have finite-dimensional sufficient statistics, regardless of the number of examples presented \cite{Fisher1934,Darmois1935,Pitman1936,Koopman1936}.
As a result, exponential family posteriors allow efficient sequential Bayesian updates, yielding results identical to offline learning \cite{Lee2024LearningTC}.
Nonetheless, their limited representational capacity prevents them from being applied to complex high-dimensional domains, where deep neural networks excel.

MOL and MCL settings provide a unique solution to this problem, which has not been possible in non-meta-learning settings: combining meta-learned neural networks and simple statistical models with exponential family posteriors.
Within this framework, only the statistical model is responsible for online or continual learning, while the meta-learned neural networks function as a bridge between the complex data space and the latent space for the statistical model.
Note that the neural network components are protected from learning issues (e.g., forgetting) in the inner loop since they are updated only in the outer loop.
In summary, this approach combines the best of both worlds: the representational capacity of neural networks and the online/continual learning ability of simple statistical models.

\paragraph*{PN \cite{Snell2017PrototypicalNF} and GeMCL \cite{Banayeeanzade2021GenerativeVD}.}
While introduced as a metric-based meta-learning method for classification in \S\ref{sec:prelim:ml:metric}, Prototypical Network (PN) \cite{Snell2017PrototypicalNF} can be reinterpreted as a combination of a neural network encoder and a Gaussian mixture model (GMM) with an isotropic Gaussian component for each class \cite{Lee2024LearningTC}.
In this context, class prediction becomes inferring which GMM component is responsible for the embedding produced by the encoder.
If we assume an uninformative prior for each Gaussian component of the GMM, computing the posterior mean of each Gaussian reduces to calculating the average of the embeddings belonging to the corresponding class.
This process was originally called computing the prototypes in \cite{Snell2017PrototypicalNF}.
Since the averages can be computed incrementally as more data becomes available, PN can be applied to MOL or MCL settings immediately \cite{Banayeeanzade2021GenerativeVD}.
Generative Meta-Continual Learning (GeMCL) \cite{Banayeeanzade2021GenerativeVD} is a generalization of PN that employs factorized Gaussian components and meta-learned priors for the GMM.
Since the posterior of each component is in the form of a normal-gamma distribution, there is a closed-form sequential Bayesian update rule for GeMCL.

\paragraph*{ALPaCA \cite{Harrison2018MetaLearningPF}.}
While the choice of GMM makes PN and GeMCL suitable for classification tasks, ALPaCA \cite{Harrison2018MetaLearningPF} employs a linear model for regression problems.
The posterior of the linear model's weights follows the matrix-normal distribution, and the linear model formulation also features an efficient sequential Bayesian update rule for the posterior.

\paragraph*{}
PN, GeMCL, and ALPaCA all share a similar model architecture: a neural network encoder followed by a simple statistical model.
The encoder first processes a complex high-dimensional input into a streamlined embedding, which is subsequently fed into the statistical model to produce the final output.
However, since the statistical models directly produce the final output, it is hard to apply these methods to other domains that require more complicated outputs.
In addition, they cannot handle unsupervised learning scenarios, as they are specifically designed for classification or regression problems.

\begin{figure}
  \includegraphics[width=\linewidth]{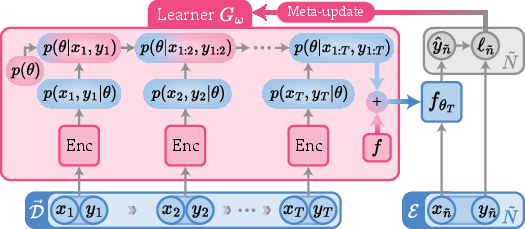}
  \caption{
    Learning as sequential Bayesian update (MCL setting).
  }
  \label{fig:mcl:bayes}
\end{figure}

\paragraph*{SB-MCL \cite{Lee2024LearningTC}.}
Sequential Bayesian Meta-Continual Learning (SB-MCL) \cite{Lee2024LearningTC} was proposed as a general MCL framework that can handle arbitrary domains.
Fig.~\ref{fig:mcl:bayes} illustrates the overall framework, which also covers PN, GeMCL, and ALPaCA as special cases.
Within this framework, the model is primarily divided into two parts: a simple statistical model with parameter $\modelparam$ and other meta-learned neural components.
In the case of PN or GeMCL, $\modelparam$ corresponds to the GMM's parameters, and in ALPaCA, it refers to the parameter of the linear model.
For general cases, \cite{Lee2024LearningTC} employs a factorized Gaussian variable.
When a new example $\exn{t+1}$ is provided in the inner loop, a meta-learned encoder processes both $x_{t+1}$ and $y_{t+1}$ into the likelihood $p(x_{t+1}, y_{t+1} | \modelparam)$, which is subsequently combined into the posterior (or its variational approximation) with the Bayes rule, i.e., $p(\modelparam | x_{1:t+1}, y_{1:t+1}) \propto p(\modelparam | x_{1:t}, y_{1:t}) p(x_{t+1}, y_{t+1} | \theta)$.
At test time, the final output is computed by marginalizing out $\modelparam$ analytically, 
i.e., $\int_\modelparam p(\modelparam | x_{1:T}, y_{1:T}) \model_{\modelparam}(x_{\tilde n})$,
or adopting approximation techniques for $\modelparam$, such as Monte Carlo or maximum a posteriori estimations.
This abstract framework allows a wide range of models to be adapted for MCL or MOL: one can modify an existing model to take additional input $\modelparam$ while meta-learning other parameters.
\cite{Lee2024LearningTC} demonstrates the effectiveness of SB-MCL in supervised settings with complex output space, such as image completion, and also in unsupervised settings by adapting deep generative models such as variational autoencoder \cite{Kingma2013AutoEncodingVB} and diffusion models \cite{Ho2020DenoisingDP}.

\subsection{Learning as Sequence Modeling}
\label{sec:mcl-mol:seq}

The approaches previously introduced in \S\ref{sec:mcl-mol:sgd} and \S\ref{sec:mcl-mol:bayes} impose a specific structure on the learning rule, i.e., SGD and sequential Bayes.
Although such an explicit structure can be beneficial when the problem domain is compatible with it, this may not always be the case.
For instance, the sequential Bayesian approaches in \S\ref{sec:mcl-mol:bayes} produce the same learning outcome regardless of the order of the training data since the posterior belongs to the exponential family.
This property can be beneficial if the order of training examples is irrelevant, but otherwise, when there is a strong dependency among examples based on their order, it may impede learning.
In such scenarios, minimizing the structural prior on the learning rule can be a solution.

\begin{figure}
  \includegraphics[width=\linewidth]{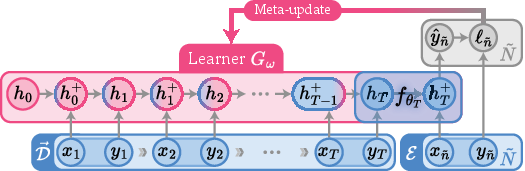}
  \caption{
    Learning as sequence modeling (MCL setting).
  }
  \label{fig:mcl:seq}
\end{figure}

In this line of research, \cite{lee2023recasting} pointed out that CL is inherently a sequence modeling problem.
It showed that the CL objective can be represented as the autoregressive objective in the sequence modeling literature.
Specifically, the goal of continual learning is to predict a test target $y_{\tilde n}$ given a test input $x_{\tilde n}$ after learning a training stream $\trainstreamdef$.
This can be expressed as predicting the next token $y_{\tilde n}$ that comes after the sequence $x_1, y_1, ..., x_T, y_T, x_{\tilde n}$, as depicted in Fig.~\ref{fig:mcl:seq}.
From this perspective, the CL process becomes the forward pass of a recurrent sequence model; for each training example, learning occurs as the sequence model's internal state $h_t$ is updated by the forward pass.
The sequence model conditioned on $h_t$ can be thought of as a virtual model $\model_{\modelparam_t}$ that is sequentially updated as more training examples are provided.
In the language modeling literature, this learning mechanism is referred to as \emph{in-context learning} \cite{Brown2020LanguageMA}.
By adopting the MCL setting, the sequence model can be trained at the meta-level on multiple CL episodes.
Within this framework, the overall MCL scheme becomes equivalent to standard sequence model training.
Therefore, \cite{lee2023recasting} argues that technically, any autoregressive sequence model can be applied as a solution to MCL.
As examples, this work demonstrated that Transformers \cite{Vaswani2017AttentionIA} and efficient Transformers \cite{Katharopoulos2020TransformersAR,Choromanski2020RethinkingAW,Tay20M23Efficient} show strong MCL performances.
This framework can easily be adapted to MOL settings by making prediction of $y_{t+1}$ given $h_t$ and $x_{t+1}$ during online training.

In the meta-learning literature, some model-based approaches are aligned with this direction.
For example, \cite{Ravi2016OptimizationAA} and \cite{Santoro2016MetaLearningWM} proposed RNN-based architectures, and \cite{Mishra2017ASN} introduced an architecture that combines temporal convolution and attention layers.
Since these approaches can also be classified as autoregressive sequence models, they can be applied to MOL and MCL settings.
In the meta-reinforcement learning literature, \cite{Duan2016RL2FR, Nagabandi2018LearningTA, Lu2023StructuredSS} proposed meta-training sequence models to perform RL as the in-context learning of a sequence model.
Since RL naturally involves some form of online learning, these approaches also fall into the MOL or MCL category.

Adopting sequence models for online or continual learning offers several distinct advantages.
Firstly, this framework imposes minimal constraints on the learning dynamics, as the learning process is equivalent to the forward pass of a sequence model.
This potentially enables acquiring more flexible update rules, compared to SGD or sequential Bayesian updates.
Furthermore, numerous advancements within the sequence modeling domain can be leveraged for MOL or MCL problems.
However, some limitations in existing sequence models must be addressed before applying them to more practical applications.
For instance, current sequence models struggle with handling long sequences.
To extend the application of sequence models to much longer episodes, encompassing millions of examples, there needs to be a breakthrough in sequence modeling technology.

\subsection{Comparative Analysis}
\label{sec:mcl-mol:comparison}

The three categories of MOL and MCL approaches we have discussed have their own strengths and weaknesses.
In the following, we compare these approaches in terms of performance, parallelizability, order-sensitivity, and length generalization.
For the sequence modeling approaches, we restrict our discussion to Transformers \cite{lee2023recasting} while other sequence models can have different characteristics.

\paragraph*{CL Performance.}
According to \cite{Lee2024LearningTC} and \cite{lee2023recasting}, SGD-based approaches generally perform worse than the other two categories.
Since SGD is a greedy algorithm that updates the current parameter in the steepest direction, it is relatively difficult to control the learning dynamics by adjusting the initialization point alone.
On the other hand, the sequential Bayes and sequence modeling approaches can represent the learning dynamics more flexibly, which is crucial for embedding meta-learned knowledge into the learning process.

\paragraph*{Parallelization.}
The CL process in the inner loop is inherently a sequential process.
However, if there is an equivalent parallel computation rule for the inner loop, the meta-training process can be significantly accelerated with modern parallel processors like GPUs.
When using Transformers, the CL process is a forward pass of the model, which can be parallelized.
The sequential Bayesian update of exponential family posteriors is also parallelizable, as there is an equivalent batch update rule for the sequential update.
On the other hand, multiple gradient descent steps cannot be parallelized, as the update at each step depends on the previous step.
As a result, the SGD-based approaches take several times longer to meta-train than the other two categories \cite{Lee2024LearningTC,lee2023recasting}.

\paragraph*{Order-Sensitivity.}
While the exponential family posterior provides lossless sequential Bayesian updates, it has a fundamental limitation in its representational capacity: ignoring the order of training data \cite{Lee2024LearningTC}.
The exponential family posterior is defined by the sufficient statistics of the data, which are defined to be invariant to the order of the data.
This invariance can be beneficial in some scenarios, but it can also be a limitation when the order of the data is crucial, such as curriculum learning \cite{CurriculumLearning} or reinforcement learning \cite{Sutton2018Reinforcement}.
SGD-based approaches are not invariant to the order of the data, but they are not explicitly designed to be order-sensitive.
On the other hand, autoregressive sequence models can be order-sensitive if the model is designed to be so.
For example, Transformers can be designed to be order-sensitive by incorporating positional encodings \cite{Vaswani2017AttentionIA}.
Therefore, the sequence modeling approaches can be most flexible in terms of order-sensitivity.

\paragraph*{Length Generalization.}
To handle indefinitely long CL episodes, the learning algorithm must be able to generalize to lengths that are not seen during meta-training.
The experiments of \cite{Lee2024LearningTC} show that the Bayesian approaches are the most robust to unseen lengths, thanks to the lossless sequential Bayesian update of exponential family posteriors.
On the other hand, Transformers are notoriously bad at generalizing to unseen lengths \cite{TFLength}.
Designing a Transformer that can generalize to unseen lengths is currently an active research topic \cite{ALiBi,ruoss-2023-randomized}.

\section{Online Meta-Learning \& Continual Meta-Learning}
\label{sec:oml-cml}

In this section, we review prior studies on online meta-learning (OML) and continual meta-learning (CML).
These learning frameworks aim to incrementally improve a learning algorithm from a \emph{stream} of episodes.
Similar to the previous section, we do not draw a clear boundary between OML and CML since many methods proposed in one field are compatible with the other.

The vast majority of existing OML and CML approaches are based on sequentially updating the model's initial parameter $\modelparam^0$, which is updated into an episode-specific parameter $\modelparam^u$ for each episode $\epim{u}$.
They can be considered as extensions of MAML \cite{Finn2017ModelAgnosticMF}, sharing the idea of ``meta-learning initializations.''
Therefore, we primarily cluster existing approaches by how they manage the initializations: unitary initialization, a mixture of initializations, and compositional initialization.
Meanwhile, they can also be categorized by the form of each learning episode.
In \S\ref{sec:oml-cml:epi}, we briefly introduce several examples with different types of episodes, and in subsequent sections (\S\ref{sec:oml-cml:maml}-\ref{sec:oml-cml:comp}), we delve into each type of initialization schemes.

\subsection{Categorization by the Type of Learning Episodes}
\label{sec:oml-cml:epi}

\paragraph*{Offline Learning Episodes.}
The most basic form of OML and CML involves offline learning episodes, as outlined in Alg.~\ref{alg:continual-meta}.
For each episode, a complete training set is provided and can be freely accessed during model training.
Unless specified otherwise, this is the default setting in the following subsections.

\paragraph*{Online Learning Episodes.}
Several studies have replaced the offline learning episodes in the basic OML/CML settings with online learning episodes \cite{Denevi2019OnlineWithinOnlineM, He2019TaskAC, Harrison2019ContinuousMW, Caccia2020OnlineFA, Ren2020WanderingWA}.
Since both episodes and the examples within them are presented online, this framework inherently establishes a \emph{fully online} setup, where all examples from distinct episodes form a single continuous sequence.
Notably, in several studies \cite{He2019TaskAC, Harrison2019ContinuousMW, Caccia2020OnlineFA} more challenging scenarios were addressed where the boundaries between episodes were not explicitly provided.

\paragraph*{Reinforcement Learning Episodes.}
Another sub-branch of OML/CML aims to incrementally improve reinforcement learning ability \cite{Nagabandi2018DeepOL}.

\begin{figure}
  \centering
  \begin{subfigure}[b]{0.2\linewidth}
      \centering
      \includegraphics{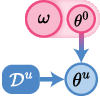}
      \caption{Unitary}
      \label{fig:oml:uni}
  \end{subfigure}
  \hfill
  \begin{subfigure}[b]{0.3\linewidth}
      \centering
      \includegraphics{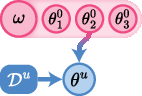}
      \caption{Mixture}
      \label{fig:oml:mixture}
  \end{subfigure}
  \hfill
  \begin{subfigure}[b]{0.4\linewidth}
      \centering
      \includegraphics{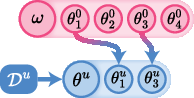}
      \caption{Compositional}
      \label{fig:oml:comp}
  \end{subfigure}
  \caption{
    Types of model initialization schemes in MAML-based OML/CML approaches.
    In MAML-based approaches, the model initialization $\modelparam^0$ is a major part of the meta-learned parameter $\learnerparam$.
    Given a training set $\trainset^u$ of episode $u$, this initialization is updated to the episode-specific model parameter $\modelparam^u$.
    (a) In unitary initialization, a single initial parameter $\modelparam^0$ is meta-learned for all types of episodes.
    (b) Mixture-based initialization scheme maintains a set of initializations $\{\modelparam_l^0\}_{l=1}^L$, and the best initialization is chosen for each episode.
    (c) In compositional initialization, a subset of the initialization components are selected to constitute the initialization for each episode.
  }
  \label{fig:oml}
\end{figure}

\subsection{Unitary Initialization}
\label{sec:oml-cml:maml}

The unitary initialization scheme is the most straightforward extension of MAML: updating a single initialization point $\modelparam^0$ for all kinds of episodes.
The model initialization is updated gradually through SGD as the learner encounters episodes sequentially.
As illustrated in Fig.~\ref{fig:oml:uni}, the same initialization $\modelparam^0$ is used regardless of the type of episode.

\paragraph*{FTML \cite{Finn2019OnlineM}.}
Follow the Meta-Leader (FTML) is a direct adaptation of Follow the Leader (FTL) approaches \cite{Hannan1957Approximation,Kalai2005Efficient} in the online learning literature.
The central concept of FTL is to maintain the best possible model for all the data received up to the current time step.
In other words, at each time step $t$, it updates the model parameter as follows: $\modelparam_t = \argmin_\modelparam \Ls(\model_\modelparam, \{ x_1, \dots, x_t \})$ where $\Ls$ is some loss function that evaluates a model on a dataset.
Note that this approach typically involves storing all incoming data in a buffer to execute the update.
FTML applies the same fundamental principle within the framework of MAML \cite{Finn2017ModelAgnosticMF}.
Once $u$ episodes have been observed, the learner's parameter $\learnerparam_u$, which is the initial model parameter in this case, is updated to minimize the meta-loss across all previously encountered tasks, i.e., $\learnerparam_u = \argmin_\learnerparam \sum_{u'=1}^u \Ls(\learner_\learnerparam(\trainset^{u'}), \testset^{u'})$.
To make this objective computationally manageable, FTML employs stochastic gradient descent with episodes uniformly sampled from the set of $u$ episodes that have been provided up to that point.
Since all previous episodes are kept in memory, FTML can be regarded as a replay-based CL approach with unlimited replay buffer.

\paragraph*{MOML \cite{Acar2021MemoryEO}.}
Memory-Efficient Online Meta-Learning (MOML) aims to alleviate the necessity of storing all previous episodes in \cite{Finn2019OnlineM}.
Its main idea is to introduce two regularization terms.
The first term adds the meta-gradient from the previous episode to the meta-gradient of the current episode, which is expected to have a similar effect as momentum SGD.
The other term maintains an exponential moving average of the meta-parameter and encourages the updated meta-parameter to stay close to it by imposing L2 regularization between the two.
Since it only keeps the meta-gradient from the previous episode and the exponential moving average, each of which has the same size as the model parameter, it can save memory consumption compared to storing the entire history of episodes.

\paragraph*{BOML \cite{Yap2020AddressingCF}.}
Bayesian Online Meta-Learning (BOML) combines Bayesian online learning (BOL) \cite{871c6579c7e14fa19571ab1b4e6f6f22} with MAML.
Employing the recursive posterior update rule of BOL framework, BOML iteratively updates the posterior of parameter as it sequentially encounters distinct few-shot classification episodes.
Since obtaining an exact posterior is typically intractable for entire neural network, the authors employ two approximation techniques adapted to address distribution shift: Laplace approximation \cite{Ritter2018OnlineSL} and variational inference \cite{Nguyen2017VariationalCL}.

Furthermore, the unitary initialization scheme has been explored within fully online setups \cite{He2019TaskAC, Caccia2020OnlineFA}, which involve online learning episodes.

\subsection{Mixture of Initializations}
\label{sec:oml-cml:mixture}

In contrast to the single initialization point in the unitary initialization $\modelparam^0$, there is another approach that maintains a set of initializations $\{\modelparam_l^0\}_{l=1}^L$.
The key assumption of these approaches is that the episodes can be clustered into $L$ groups with distinct characteristics.
In such cases, it can be beneficial to have multiple initialization points, each of which is specialized for a group of episodes.
When a new episode appears, a suitable initialization is chosen for the episode.
This setting can be formalized into the mixture model framework \cite{McLachlan2000Mixture}, where each initialization corresponds to a mixture component.
Therefore, we call this approach \emph{mixture of initializations}.

\paragraph*{Dirichlet Process Mixture of Initializations \cite{Jerfel2018ReconcilingMA,Nagabandi2018DeepOL}.}
A major challenge in the mixture model formulation is to determine the number of mixture components.
One way is to simply set it as a hyperparameter and find a proper value through a hyperparameter search.
However, this may not be a feasible option, especially when the optimal number of components can vary widely.
Dirichlet process mixture model (DPMM) \cite{Antoniak1974MixturesOD,FERGUSON1983287} is a nonparametric Bayesian framework that can be a solution to this problem.
Instead of the number of components, the hyperparameter of DPMM is a positive real value called \emph{the concentration parameter}, which controls how easily a new component is added.
With an appropriate concentration parameter, DPMM can find the number of components suitable for given data.
\cite{Jerfel2018ReconcilingMA} brings this DPMM framework into CML settings with a sequential approximation technique similar to \cite{Lin2013OnlineLO}.
\cite{Nagabandi2018DeepOL} applies a similar idea in model-based RL scenarios, where an agent should quickly adapt to new environment dynamics.
As a result, they can adaptively control the number of initializations as it encounters new episodes online.

\paragraph*{VC-BML \cite{Zhang2021VariationalCB}.}
Variational Continual Bayesian Meta-Learning (VC-BML) extends this online mixture strategy in two aspects.
First, the authors challenged the assumption made in \cite{Jerfel2018ReconcilingMA}, wherein $\theta^{(\cdot)}$ is treated as a $\delta$ distribution.
Instead, they adopt a Gaussian mixture model to represent the distributions of $\theta^{(\cdot)}$.
Second, the authors addressed the constraint of relying on MAP for inference.
To overcome this limitation, they proposed to apply a structured variational inference approach to the posterior.

\subsection{Compositional Initialization}
\label{sec:oml-cml:comp}

In a mixture of initializations, an episode is handled by a single mixture component.
On the other hand, compositional initialization combines a subset of components for each episode.
This approach can produce combinatorially more diverse initializations within the same memory budget.
Moreover, since the same component can be reused in various combinations, knowledge can be shared among different types of episodes.

\paragraph*{OSML \cite{Yao2020OnlineSM}.}
Online Structured Meta-Learning (OSML) maintains a set of initial modules for each layer of a model.
It chooses one module for each layer to build an episode-specific model.
To find the best combination of modules, OSML first represents each layer as a weighted sum of the modules with learnable weighting values.
Then, the entire network, including the module weightings and individual modules, is trained on the given training set.
Once the training converges, the module with the highest weighting value is selected for each layer to construct the episode-specific model architecture.
This architecture is further fine-tuned on the training set to compensate for the lack of other modules.
This final model is evaluated on the test set, and the initial modules and the initial weighting value for each module are meta-updated as in MAML \cite{Finn2017ModelAgnosticMF}.
Note that this approach follows the FTML \cite{Finn2019OnlineM} setting where previous episodes can be accessed freely, which is not a typical OML/CML setting.
Therefore, the meta-level forgetting issue can be avoided.

\paragraph*{ACML \cite{pmlr-v202-wu23d}.}
Adaptive compositional Continual Meta-Learning (ACML) is a counterpart of the DPMM-based approach of \cite{Jerfel2018ReconcilingMA} in the compositional initialization domain.
For compositional initialization with an adaptive number of components, the Beta process prior \cite{Hjort1990Beta,Griffiths2011TheIB} is employed, instead of the Dirichlet process prior for the nonparametric mixture model in \cite{Jerfel2018ReconcilingMA}.
In contrast to the mixture model where a single component is selected for an instance, the Beta process prior is better suited for nonparametric factorial scenarios \cite{Meeds2006Factor} where an arbitrary number of components (the factors) can be \emph{on} or \emph{off} for a single instance, and the total number of components can vary.
In the ACML setting, each component comprises the initial parameters of the entire model.
These components are independently trained for a given training set, and a subset of the trained components are summed up to construct the final model.

\subsection{Comparative Analysis}

The suitable initialization scheme can vary depending on various factors, such as the episode distribution, the amount of data, and the model architecture.
Sharing a single initialization point for all episodes in the unitary initialization scheme can be beneficial when all the episodes are similar to each other, as it can maximally share the knowledge among episodes.
On the other hand, the mixture of initializations and compositional initialization schemes can be more suitable for scenarios where the episodes are diverse and do not have much information worth sharing \cite{Jerfel2018ReconcilingMA,Yao2020OnlineSM}.
The unitary initialization can also be more robust when the number of episodes is small, as the mixture-based and compositional initialization schemes are more flexible, potentially leading to overfitting.
The model architecture can also affect the choice of initialization scheme.
If the model's expressiveness is limited, more complex initialization schemes may complement the small information capacity of a single initialization point.
To sum up, there is no one-size-fits-all solution, and the choice of initialization scheme should be made based on the specific characteristics of the problem.

\section{Continual Bi-Level Learning}
\label{sec:cbl}
In this section, we explore continual bi-level learning (CBL; \S\ref{sec:tax:cbl}) approaches, which incorporate meta-learning methods into the CL framework.
Typically, this integration begins with the idea that extracting shared knowledge among tasks is beneficial for sequentially learning different tasks without forgetting.
We categorize the proposed CBL methods based on the employed meta-learning techniques and the specific contexts in which they are applied.
Specifically, we investigate approaches that use bi-level optimization techniques (\S\ref{sec:cbl:opt}) and hypernetworks (\S\ref{sec:cbl:hyp}).
Afterward, we review approaches on a CBL variant which regards the first task as a pseudo-meta-training set (\S\ref{sec:cbl:app}).

It is worth noting that the majority of the works discussed in this section utilize additional memory buffers to enhance the process of meta-learning.
These memory buffers serve various purposes, including the storage of encountered examples, neural networks for generative replay, and task-specific representations. 
Consequently, we will conduct a comprehensive analysis of how the incorporation of memory buffers is harmonized with each meta-learning method.

\subsection{Approaches with Bi-Level Optimization}
\label{sec:cbl:opt}

There have been various CBL approaches that incorporate bi-level optimization.
One possible strategy is to leverage bi-level optimization as a means of achieving the concept of \emph{gradient alignment} introduced in GEM \cite{LopezPaz2017GradientEM}.
As discussed in \S\ref{sec:prelim:cl}, GEM is designed to circumvent negative dot-products between gradients w.r.t. different tasks during SGD updates.
Building upon this, some approaches \cite{Riemer2018LearningTL, Gupta2020LaMAMLLM} employs MAML \cite{Finn2017ModelAgnosticMF} or Reptile \cite{Nichol2018OnFM} to align gradients as well.

\paragraph*{MER \cite{Riemer2018LearningTL}.}
Instead of merely projecting gradients in a safe direction as in prior works \cite{LopezPaz2017GradientEM, Chaudhry2018EfficientLL}, Meta-Experience Replay (MER) aims to optimize model parameters to align gradients for upcoming tasks, thereby facilitating positive knowledge transfer.
For each example within a task, MER adapts the parameter through batches constructed from past examples randomly drawn from the replay buffer, as well as the specific example itself.
Then, the parameter is updated by the Reptile update rule, obviating the need for computing the second-order derivatives.

\paragraph*{La-MAML \cite{Gupta2020LaMAMLLM}.}
In contrast, Look-ahead MAML (La-MAML) extends MAML for CBL.
When encountering a task, this method adapts the parameter with the specific task.
Subsequently, the adapted parameter is evaluated on batches constructed from examples of both current and past tasks, leveraging the replay buffer.
This evaluation process is then backpropagated to update the parameter.
To avoid the need for second-order derivative computation, La-MAML incorporates a first-order approximation technique.
Additionally, this method optimizes the learning rates of inner loop on a per-parameter basis.
This approach was further extended in \cite{Oswald2021LearningWT}.

Furthermore, there have been distinct lines of research, also tackling CL with bi-level optimization strategies.

\paragraph*{iTAML \cite{Rajasegaran2020iTAMLAI}.}
A key distinction of Incremental Task-Agnostic Meta-learning (iTAML) from aforementioned approaches \cite{Riemer2018LearningTL, Gupta2020LaMAMLLM} is that it allows task-specific test time adaptation by inferring the task.
Similar to \cite{Riemer2018LearningTL}, iTAML updates the parameter using the Reptile algorithm and a replay buffer.
During the test time, iTAML infers the task identity and updates the parameter using the examples stored in the replay buffer, which belong to the specific task.
Subsequently, it proceeds to make class predictions for the test example.

\paragraph*{Continual Domain Adaptation.}
\cite{Volpi2020ContinualAO} focused on continual domain adaptation, which involves presenting slightly varying visual recognition benchmarks in a sequential manner.
For example, this method aims to train a model sequentially on digit recognition benchmarks \cite{LeCun1998GradientbasedLA, Netzer2011ReadingDI, Ganin2014UnsupervisedDA, Ganin2014UnsupervisedDA} without catastrophic forgetting.
In this context, the term \emph{domain} refers to each slightly distinct task.
What distinguishes the approach of \cite{Volpi2020ContinualAO} from others is the use of predetermined image transformation functions, instead of a replay buffer.
Upon encountering a new task, these functions are respectively applied to the task, creating multiple transformed tasks.
These transformed tasks then act as the past tasks stored in the replay buffer in the method of \cite{Gupta2020LaMAMLLM}.

As mentioned in \S\ref{sec:prelim:ml:opt}, the straightforward implementation of bi-level optimization can pose significant computational challenges.
Therefore, opting for an appropriate approximation technique can be beneficial, as demonstrated in introduced works \cite{Riemer2018LearningTL, Gupta2020LaMAMLLM, Rajasegaran2020iTAMLAI}.

\subsection{Approaches with Hypernetworks}
\label{sec:cbl:hyp}
Hypernetworks, in the context of neural networks, refer to networks that are specifically designed to generate the parameters of other networks \cite{Ha2016HyperNetworks, Jia2016DynamicFN, Krueger2017BayesianH, Savarese2019LearningIR}.
Enabling dynamic model generation during the test time, these hypernetworks have found application in various domains of deep learning, including continual learning \cite{Oswald2019ContinualLW, Hu2018OvercomingCF, Joseph2020MetaConsolidationFC, Ehret2020ContinualLI, Chandra2022ContinualLW, Hemati2023PartialHF} and meta-learning \cite{Bertinetto2016LearningFO, Munkhdalai2017MetaN, Zhao2020MetaLearningVH, Zhmoginov2022HyperTransformerMG}.
When hypernetworks are incorporated into the CL framework, it falls under our definition of CBL (\S\ref{sec:tax:cbl}).
The precise mechanism through which a hypernetwork generates a network varies across different works.

\paragraph*{Task-Conditioned Hypernetworks \cite{Oswald2019ContinualLW}.}
One straightforward strategy for incorporating a hypernetwork into CL involves generating task-specific models by providing task identity during the test.
For example, in \cite{Oswald2019ContinualLW}, a hypernetwork is employed to map a task embedding onto task-specific network parameters.
The embeddings are learned with the hypernetwork during training.
Additionally, they addressed catastrophic forgetting of the hypernetwork by applying a regularization (\S\ref{sec:prelim:cl:reg}) and a generative replay (\S\ref{sec:prelim:cl:replay}) techniques.
Subsequent works extended this approach further \cite{Ehret2020ContinualLI, Chandra2022ContinualLW, Hemati2023PartialHF}.

However, the naive implementation of this approach relies on a highly task-aware setup, demanding a task identity even in the test phase.
To overcome this limitation, \cite{Oswald2019ContinualLW} suggested several methods for inferring the task.
Likewise, an alternative strategy is to utilize hypernetwork to output task-specific model with a single test example as an input, inferring the task membership autonomously.

\paragraph*{PGMA \cite{Hu2018OvercomingCF}.}
For example, Parameter Generation and Model Adaptation (PGMA) employs a hypernetwork to generate task-specific parameters for each example.
This approach additionally incorporates an Wasserstein auto-encoder (WAE) \cite{Tolstikhin2017WassersteinA} for generative replay \cite{Shin2017ContinualLW}.
The role of the WAE's encoder is to map examples into a lower-dimensional space, whereas its decoder generates synthetic examples of past tasks.
The training of the hypernetwork and the WAE occurs in an alternating fashion.
During training the hypernetwork, the decoder of WAE generates synthetic examples to counteract catastrophic forgetting of the hypernetwork.

\paragraph*{MERLIN \cite{Joseph2020MetaConsolidationFC}.}
Alternatively, in Meta-Consolidation for Continual Learning (MERLIN), a variational auto-encoder (VAE) \cite{Kingma2013AutoEncodingVB} is employed as a hypernetwork itself, enabling the sampling of task-specific model parameters.
Assuming an offline CL setup, when encountering a new task, MERLIN resamples $B$ subsets of examples to train $B$ distinct model parameters. 
These parameters are then used to train a VAE, allowing it to learn the distribution of task-specific model parameters.
To mitigate catastrophic forgetting and consolidate the learned distributions of task-specific parameters, MERLIN samples model parameters of past tasks using the decoder of VAE and combines them with trained parameters for the current task.
When sampling a task-specific model parameter, a latent variable $z$ is passed to the VAE decoder, sampled from a Gaussian distribution $\mathcal{N}(z|\mu_t, \Sigma_t)$, where ($\mu_t, \Sigma_t$) represents the \emph{task prior} for task $t$, which is stored in a replay buffer.
During the test, MERLIN samples several task-specific parameters and ensembles them to make predictions.
To avoid the need for task identity during the test, the authors explored an additional setup where the \emph{average} of all task priors is used to sample $z$ for all tasks.

As evident from our review, the challenge of training hypernetworks while avoiding catastrophic forgetting has consistently captured the attention of prior research efforts.
It is noteworthy that attempts to address this challenge have frequently employed regularization (\S\ref{sec:prelim:cl:reg}) and generative replay (\S\ref{sec:prelim:cl:replay}) methods.

\subsection{First Task as a Pseudo-Meta-Training Set}
\label{sec:cbl:app}
Previous studies have modified CL frameworks to better suit practical learning scenarios.
An example of such case involves incorporating a preliminary offline learning phase before initiating continual learning.
During the offline learning phase, an abundance of examples from \emph{base} task is provided, whereas in the continual learning phase, only a limited number of examples from \emph{novel} task are available.
The model is required to incrementally learn the novel task without revisiting the base task, while maintaining performance on all previously encountered classes.

In this setup, meta-learning methods can be effectively leveraged to extract knowledge during the offline learning phase, which is then applied during the continual learning phase.
In other words, the first task is utilized as a \emph{pseudo}-meta-training set.
This concept has been explored across several application domains.

\paragraph*{FSCIL \cite{Tao2020FewShotCL}.}
Few-Shot Class-Incremental Learning (FSCIL) is a such variant of CL designed for image classification \cite{Tao2020FewShotCL, Cheraghian2021SemanticawareKD, Dong2021FewShotCL, Zhou2022ForwardCF, Cheraghian2021SynthesizedFB, Mazumder2021FewShotLL, Hersche2022ConstrainedFC, Zhu2021SelfPromotedPR, Zhang2021FewShotIL, Zhu2023FewshotIL, Zhao2023FewShotCL, Zhuang2023GKEALGK}.
Some studies have investigated simulating the continual learning phase within the offline learning phase and have adopted meta-learning techniques to replicate meta-learning processes \cite{Chi2022MetaFSCILAM, Zhou2022FewShotCL}.
For instance, MetaFSCIL \cite{Chi2022MetaFSCILAM} trains a network using a modulation mechanism \cite{Beaulieu2020LearningTC}, incorporating a bi-level optimization technique.
Similarly, in \cite{Zhou2022FewShotCL}, a Transformer \cite{Vaswani2017AttentionIA} is trained during the offline learning phase to calibrate the embedding of novel class examples with the old classifier.

\paragraph*{iFSD \cite{PrezRa2020IncrementalFO}.} 
Another comparable variant of CL is Incremental Few-Shot Object Detection (iFSD) \cite{PrezRa2020IncrementalFO, Li2020IncrementalFO, Dong2022IncrementalDETRIF, Cheng2022MetaLearningBasedIF, Yin2022SylphAH, Choi2023IncrementalFO}, a modification of incremental object detection \cite{Shmelkov2017IncrementalLO, Feng2022OvercomingCF, Joseph2020IncrementalOD}.
In this scenario, base classes are abundantly annotated, while novel classes are sparsely annotated.
One notable approach was presented by \cite{PrezRa2020IncrementalFO}, which enhances CenterNet \cite{Zhou2019ObjectsAP} by training a hypernetwork in the offline learning phase.
Similarly, Sylph \cite{Yin2022SylphAH} utilizes FCOS \cite{Tian2019FCOSFC} as the object detector and employs a hypernetwork responsible for generating classifiers weights.

In specific practical applications, it would be reasonable for a company to initially train a model using a substantial dataset prior to its deployment.
After deployment, the model might require further training with a smaller dataset in an online fashion.
FSCIL and iFSD align with this learning scenario.
These learning frameworks can also be seen as extensions of prior works \cite{Qiao2017FewShotIR, Gidaris2018DynamicFV, Ren2018IncrementalFL}.

\section{Practical Applications}
\label{sec:app}

In this section, we explore the practical applications of research within our scope, highlighting how meta-learning, continual learning, and online learning intersect and complement each other in various domains.
Since some applications do not fit neatly into a single learning framework, we present a broad range of examples that combine elements of these learning frameworks.

\paragraph*{Robot Manipulation.}
The integration of different learning frameworks is particularly useful in robot manipulation, where agents must adapt to a variety of environments and tasks over time.
For instance, \cite{Nagabandi2018DeepOL} and \cite{Nagabandi2018LearningTA} applied meta-learning to adapt the dynamics model within the context of model-based RL.
These authors demonstrated their approach by having robotic agents navigate new terrains and adapt to physical impairments.
Other studies, such as \cite{Bing2022MetaReinforcementLI}, have focused on agents that continuously adjust to non-stationary environments, especially in competitive multi-agent settings \cite{AlShedivat2017ContinuousAV, Kim2020APG}.
More recently, \cite{Lu2023StructuredSS} employed a structured state-space model \cite{Gu2021EfficientlyML} within the in-context RL framework, further enhancing the adaptive and generalizing capabilities in robotic systems.

\paragraph*{Large Language Models.}
Large language models (LLMs) can experience knowledge depreciation, necessitating solutions like online or continual fine-tuning to maintain their relevance.
Numerous studies have developed benchmarks to assess the performance of LLMs in temporal adaptation \cite{Lazaridou2021MindTG, Jang2021TowardsCK, Jang2022TemporalWikiAL, Livska2022StreamingQAAB, Kim2023CarpeDO}.
Some works specifically utilize meta-learning methods; for example, \cite{Dhingra2021TimeAwareLM} proposed modeling the joint distribution of text and timestamps, while \cite{Clark2022MetaLearningFW} and \cite{Hu2023MetaLearningOA} suggested meta-learning approaches for layer fine-tuning and loss modulation during online updates, respectively.
This research direction generally falls under OML/CML or CBL.
Additionally, CBL of LLMs is beneficial for various downstream natural language tasks \cite{Obamuyide2019MetaLearningIL, Wang2020EfficientML, Wu2021CurriculumMetaLF, Jin2021LearnCG}.

\section{Challenges \& Future Directions}
\label{sec:future}
Although there have been numerous research efforts at the intersection of meta-learning, online learning, and continual learning, most of the existing works are still in their nascent stages.
In this section, we discuss the challenges and future directions for real-world applications of these learning frameworks.

\subsection{Meta-Online Learning \& Meta-Continual Learning}

\paragraph*{Data Collection.}
Collecting data for meta-learning is a challenging task, as it requires a large number of learning episodes.
As a result, the current benchmarks in MOL and MCL are typically constructed by repurposing existing datasets \cite{Javed2019MetaLearningRF,Lee2024LearningTC,lee2023recasting}, which are not specifically designed for these learning frameworks.
However, to apply MOL and MCL to real-world applications, it is essential to gather meta-training data that reflects the target domain.

\paragraph*{Other Learning Problems.}
The MOL and MCL frameworks can be combined with various learning problems that can also benefit from meta-learning, such as few-shot learning \cite{NIPS2016_90e13578,Snell2017PrototypicalNF}, reinforcement learning \cite{Finn2017ModelAgnosticMF}, imitation learning \cite{ImitationLearningSurvey}, and active learning \cite{ActiveLearningSurvey}.
Realistic applications often involve a combination of these learning problems, and developing methods that can handle such complex scenarios is a promising direction for future research.

\paragraph*{Sequence Models.}
From the perspective of \S\ref{sec:mcl-mol:seq}, continual learning can be addressed using the in-context learning ability of sequence models.
Given the generality of this framework and the significant success of sequence models in various tasks, this approach holds great promise.
Nevertheless, current sequence models like Transformer face challenges in handling longer sequences.
For example, the Transformer model has a quadratic complexity in relation to sequence length.
Although multiple attempts have been made to address this issue \cite{Tay20M23Efficient}, they suffer from non-trivial performance degradation.
As we explained in \S\ref{sec:mcl-mol:comparison}, exponential-family posteriors are the only way to achieve lossless CL with constant memory, but they completely ignore the ordering of the data \cite{Lee2024LearningTC}.
Therefore, although many sequence models have been proposed to maintain the hidden state size constant, it may be necessary to expand the model's memory as it encounters new data.
Another challenge is length generalization.
To handle indefinitely long CL episodes, the model must be able to handle sequences with arbitrary lengths that are not seen during meta-training.
However, Transformers are known to have difficulty generalizing to unseen lengths \cite{TFLength}.
Especially, common positional encodings, such as the sinusoidal encoding \cite{Vaswani2017AttentionIA} or the rotary positional embedding \cite{RoPE} are not designed to generalize to unseen lengths.
Developing a generalizable positional encoding scheme is currently an active area of research \cite{LengthExtrapolationPE}.

\subsection{Online Meta-Learning \& Continual Meta-Learning}

\paragraph*{Beyond Meta-Learning Initializations.}
The vast majority of OML and CML methods focus on learning initializations for the model parameters.
They are essentially sequential variations of MAML \cite{Finn2017ModelAgnosticMF} or Reptile \cite{Nichol2018OnFM}, which are gradient-based meta-learning methods.
However, as summarized in \S\ref{sec:prelim:ml}, there are other branches of meta-learning methods, such as model-based meta-learning (\S\ref{sec:prelim:ml:model}) and metric-based meta-learning (\S\ref{sec:prelim:ml:metric}).
Exploring these other branches of meta-learning within the OML and CML frameworks is a promising direction for future research.

\paragraph*{Additional Meta-Learning for OML and CML.}
The outer loop of OML and CML is typically implemented as a gradient-based optimization process over a stream of episodes.
However, MOL and MCL approaches in \S\ref{sec:mcl-mol} show that gradient descent may not be the ideal learning mechanism for streaming data.
Since MOL and MCL have meta-learning in the outer-most loop, we can adopt more flexible learning mechanisms in the inner loop as in \S\ref{sec:mcl-mol:bayes} and \S\ref{sec:mcl-mol:seq}.
Likewise, if we wrap the OML and CML frameworks with an additional meta-learning loop, we can potentially improve the performance of these methods in a data-driven manner while using more efficient learning mechanisms for OML and CML.
The resulting framework can be considered as a triple-loop learning framework, meta-online-meta-learning or meta-continual-meta-learning.
This approach may seem computationally expensive.
However, similar to the argument in \S\ref{sec:tax:mcl-mol}, human evolution can be an example of such a triple-loop learning system;
individual lifetime is a double-loop OML or CML, and the evolution is the outer-most loop for optimizing the OML or CML ability.
Therefore, with the right design, the triple-loop learning framework can be computationally feasible and an interesting research direction.

\subsection{Continual Bi-Level Learning}

The challenges and open questions of CBL essentially mirror those of CL.
Therefore, we direct our readers to previous CL surveys for general remarks \cite{Parisi2018ContinualLL,DeLange2019ACL,wang2023comprehensive}.
We also provide a succinct discussion pertaining to CBL.

\paragraph*{Replay Buffer.}
In \S\ref{sec:cbl}, it is noted that the majority of CBL methods make use of an external memory resource, which is employed in various ways.
Some approaches involve directly storing a subset of examples \cite{Riemer2018LearningTL, Gupta2020LaMAMLLM, Rajasegaran2020iTAMLAI}, while others use this memory to train generative models for synthesizing examples \cite{Oswald2019ContinualLW, Hu2018OvercomingCF}.
However, as the number of tasks increases, replay-based methods generally demand a larger memory resource.
Therefore, it is crucial to address the issue of potentially impractical memory usage in replay methods.
Therefore, it is crucial for future works to mitigate reliance on external memory.

\paragraph*{Task-Awareness.}
As stated in \S\ref{sec:tax:cl}, the practicality of a CL method is greatly affected by the need for task identity during training and testing.
However, integrating certain meta-learning techniques into the CL framework without caution could introduce further demands for task identity.
For example, enabling a model to adapt at test time for specific tasks without explicit task inference would necessitate a highly task-aware setup even during testing, which in turn could restrict the applicability of the approach.
Hence, finding ways to address catastrophic forgetting without relying on task-awareness remains a significant and promising area of research.

\section{Conclusion}
\label{sec:conclusion}

In recent times, there has been a notable surge of research at the confluence of meta-learning, online learning, and continual learning; however, a reliable taxonomy and comprehensive review are conspicuously absent.
In this paper, we clarified the blurred boundaries among these perplexing learning frameworks by providing a comprehensive taxonomy and review.
To achieve this, we presented formal definition for each problem and examine the meta-learning and continual learning literature.
Building upon this foundation, we conducted a review of previous works that address meta-continual learning, continual meta-learning, and continual bi-level learning.
We hope that this paper dispels the confusion, thereby fostering the advancement of further research in these evolving domains.

\ifCLASSOPTIONcompsoc
  \section*{Acknowledgments}
\else
  \section*{Acknowledgment}
\fi

This work was supported by
the Center for Applied Research in Artificial Intelligence (CARAI) grant funded by Defense Acquisition Program Administration (DAPA) and Agency for Defense Development (ADD) (UD230017TD),
the Institute of Information \& Communications Technology Planning \& Evaluation (IITP) grants funded by the Korea government (MSIT) (No.~RS-2019-II191082, SW StarLab; No.~RS-2022-II220156, Fundamental research on continual meta-learning for quality enhancement of casual videos and their 3D metaverse transformation; No.~RS-2021-II211343, Artificial Intelligence Graduate School Program (Seoul National University)),
and the National Research Foundation of Korea (NRF) grant funded by the Korea government (MSIT) (No.~2023R1A2C2005573).

\ifCLASSOPTIONcaptionsoff
  \newpage
\fi

\bibliographystyle{IEEEtran}
\bibliography{clml-survey}

\begin{thebibliography}{100}
\providecommand{\url}[1]{#1}
\csname url@samestyle\endcsname
\providecommand{\newblock}{\relax}
\providecommand{\bibinfo}[2]{#2}
\providecommand{\BIBentrySTDinterwordspacing}{\spaceskip=0pt\relax}
\providecommand{\BIBentryALTinterwordstretchfactor}{4}
\providecommand{\BIBentryALTinterwordspacing}{\spaceskip=\fontdimen2\font plus
\BIBentryALTinterwordstretchfactor\fontdimen3\font minus \fontdimen4\font\relax}
\providecommand{\BIBforeignlanguage}[2]{{%
\expandafter\ifx\csname l@#1\endcsname\relax
\typeout{** WARNING: IEEEtran.bst: No hyphenation pattern has been}%
\typeout{** loaded for the language `#1'. Using the pattern for}%
\typeout{** the default language instead.}%
\else
\language=\csname l@#1\endcsname
\fi
#2}}
\providecommand{\BIBdecl}{\relax}
\BIBdecl

\bibitem{Hoi2021OnlineLA}
S.~C.~H. Hoi, D.~Sahoo, J.~Lu, and P.~Zhao, ``Online learning: {A} comprehensive survey,'' \emph{Neurocomputing}, 2021.

\bibitem{Parisi2018ContinualLL}
G.~I. Parisi, R.~Kemker, J.~L. Part, C.~Kanan, and S.~Wermter, ``Continual lifelong learning with neural networks: A review,'' \emph{Neural Networks}, 2018.

\bibitem{DeLange2019ACL}
M.~D. Lange, R.~Aljundi, M.~Masana, S.~Parisot, X.~Jia, A.~Leonardis, G.~G. Slabaugh, and T.~Tuytelaars, ``A continual learning survey: Defying forgetting in classification tasks,'' \emph{IEEE Transactions on Pattern Analysis and Machine Intelligence}, 2019.

\bibitem{Hospedales2020MetaLearningIN}
T.~M. Hospedales, A.~Antoniou, P.~Micaelli, and A.~J. Storkey, ``Meta-learning in neural networks: A survey,'' \emph{IEEE Transactions on Pattern Analysis and Machine Intelligence}, 2020.

\bibitem{Snell2017PrototypicalNF}
J.~Snell, K.~Swersky, and R.~S. Zemel, ``Prototypical networks for few-shot learning,'' in \emph{NeurIPS}, 2017.

\bibitem{Sung2017LearningTC}
F.~Sung, Y.~Yang, L.~Zhang, T.~Xiang, P.~H.~S. Torr, and T.~M. Hospedales, ``Learning to compare: Relation network for few-shot learning,'' in \emph{CVPR}, 2018.

\bibitem{Li2017MetaSGDLT}
Z.~Li, F.~Zhou, F.~Chen, and H.~Li, ``Meta-sgd: Learning to learn quickly for few shot learning,'' \emph{arXiv preprint arXiv:1707.09835}, 2017.

\bibitem{Li2020OnlineMF}
D.~Li and T.~M. Hospedales, ``Online meta-learning for multi-source and semi-supervised domain adaptation,'' in \emph{ECCV}, 2020.

\bibitem{Li2019FeatureCriticNF}
Y.~Li, Y.~Yang, W.~Zhou, and T.~M. Hospedales, ``Feature-critic networks for heterogeneous domain generalization,'' in \emph{ICML}, 2019.

\bibitem{Vettoruzzo2023AdvancesAC}
A.~Vettoruzzo, M.-R. Bouguelia, J.~Vanschoren, T.~S. R{\"o}gnvaldsson, and K.~Santosh, ``Advances and challenges in meta-learning: A technical review,'' \emph{IEEE Transactions on Pattern Analysis and Machine Intelligence}, 2023.

\bibitem{Vanschoren2019AML}
J.~Vanschoren, \emph{Automated Machine Learning}.\hskip 1em plus 0.5em minus 0.4em\relax Springer Cham, 2019, ch.~2, pp. 35--61.

\bibitem{Kingma2015AdamAM}
D.~P. Kingma and J.~Ba, ``Adam: {A} method for stochastic optimization,'' in \emph{ICLR}, 2015.

\bibitem{Goodfellow2014GenerativeAN}
I.~J. Goodfellow, J.~Pouget{-}Abadie, M.~Mirza, B.~Xu, D.~Warde{-}Farley, S.~Ozair, A.~C. Courville, and Y.~Bengio, ``Generative adversarial nets,'' in \emph{NeurIPS}, 2014.

\bibitem{Kingma2013AutoEncodingVB}
D.~P. Kingma and M.~Welling, ``Auto-encoding variational bayes,'' in \emph{ICLR}, 2014.

\bibitem{Ho2020DenoisingDP}
J.~Ho, A.~Jain, and P.~Abbeel, ``Denoising diffusion probabilistic models,'' in \emph{NeurIPS}, 2020.

\bibitem{Du2019ImplicitGA}
Y.~Du and I.~Mordatch, ``Implicit generation and modeling with energy based models,'' in \emph{NeurIPS}, 2019.

\bibitem{Zenke2017ContinualLT}
F.~Zenke, B.~Poole, and S.~Ganguli, ``Continual learning through synaptic intelligence,'' in \emph{ICML}, 2017.

\bibitem{LeCun1998GradientbasedLA}
Y.~LeCun, L.~Bottou, Y.~Bengio, and P.~Haffner, ``Gradient-based learning applied to document recognition,'' \emph{IEEE}, 1998.

\bibitem{Kirkpatrick2016OvercomingCF}
J.~Kirkpatrick, R.~Pascanu, N.~C. Rabinowitz, J.~Veness, G.~Desjardins, A.~A. Rusu, K.~Milan, J.~Quan, T.~Ramalho, A.~Grabska-Barwinska, D.~Hassabis, C.~Clopath, D.~Kumaran, and R.~Hadsell, ``Overcoming catastrophic forgetting in neural networks,'' \emph{Proceedings of the National Academy of Sciences}, 2016.

\bibitem{Schwarz2018ProgressC}
J.~Schwarz, W.~Czarnecki, J.~Luketina, A.~Grabska{-}Barwinska, Y.~W. Teh, R.~Pascanu, and R.~Hadsell, ``Progress {\&} compress: {A} scalable framework for continual learning,'' in \emph{ICML}, 2018.

\bibitem{Rusu2016ProgressiveNN}
A.~A. Rusu, N.~C. Rabinowitz, G.~Desjardins, H.~Soyer, J.~Kirkpatrick, K.~Kavukcuoglu, R.~Pascanu, and R.~Hadsell, ``Progressive neural networks,'' \emph{arXiv preprint arXiv:1606.04671}, 2016.

\bibitem{Aljundi2016ExpertGL}
R.~Aljundi, P.~Chakravarty, and T.~Tuytelaars, ``Expert gate: Lifelong learning with a network of experts,'' in \emph{CVPR}, 2017.

\bibitem{Aljundi2018TaskFreeCL}
R.~Aljundi, K.~Kelchtermans, and T.~Tuytelaars, ``Task-free continual learning,'' in \emph{CVPR}, 2019.

\bibitem{Lee2020AND}
S.~Lee, J.~Ha, D.~Zhang, and G.~Kim, ``A neural dirichlet process mixture model for task-free continual learning,'' in \emph{ICLR}, 2020.

\bibitem{LopezPaz2017GradientEM}
D.~Lopez{-}Paz and M.~Ranzato, ``Gradient episodic memory for continual learning,'' in \emph{NeurIPS}, 2017.

\bibitem{Rodrguez2018DontFT}
N.~D. Rodr{\'{\i}}guez, V.~Lomonaco, D.~Filliat, and D.~Maltoni, ``Don't forget, there is more than forgetting: new metrics for continual learning,'' \emph{arXiv preprint arXiv:1810.13166}, 2018.

\bibitem{PAC}
L.~Valiant, \emph{Probably Approximately Correct: Nature's Algorithms for Learning and Prospering in a Complex World}.\hskip 1em plus 0.5em minus 0.4em\relax Basic Books, Inc., 2013.

\bibitem{Lake2011OneSL}
B.~M. Lake, R.~Salakhutdinov, J.~Gross, and J.~B. Tenenbaum, ``One shot learning of simple visual concepts,'' \emph{Cognitive Science}, 2011.

\bibitem{He2019TaskAC}
X.~He, J.~Sygnowski, A.~Galashov, A.~A. Rusu, Y.~W. Teh, and R.~Pascanu, ``Task agnostic continual learning via meta learning,'' in \emph{4th Lifelong Machine Learning Workshop at ICML}, 2020.

\bibitem{Lee2018GradientBasedMW}
Y.~Lee and S.~Choi, ``Gradient-based meta-learning with learned layerwise metric and subspace,'' in \emph{ICML}, 2018.

\bibitem{yao2020automated}
H.~Yao, X.~Wu, Z.~Tao, Y.~Li, B.~Ding, R.~Li, and Z.~Li, ``Automated relational meta-learning,'' in \emph{ICLR}, 2020.

\bibitem{Andrychowicz2016LearningTL}
M.~Andrychowicz, M.~Denil, S.~G. Colmenarejo, M.~W. Hoffman, D.~Pfau, T.~Schaul, and N.~de~Freitas, ``Learning to learn by gradient descent by gradient descent,'' in \emph{NeurIPS}, 2016.

\bibitem{li2017learning}
K.~Li and J.~Malik, ``Learning to optimize,'' in \emph{ICLR}, 2017.

\bibitem{Finn2017ModelAgnosticMF}
C.~Finn, P.~Abbeel, and S.~Levine, ``Model-agnostic meta-learning for fast adaptation of deep networks,'' in \emph{ICML}, 2017.

\bibitem{Antoniou2018HowTT}
A.~Antoniou, H.~Edwards, and A.~J. Storkey, ``How to train your {MAML},'' in \emph{ICLR}, 2019.

\bibitem{Rajeswaran2019MetaLearningWI}
A.~Rajeswaran, C.~Finn, S.~M. Kakade, and S.~Levine, ``Meta-learning with implicit gradients,'' in \emph{NeurIPS}, 2019.

\bibitem{Flennerhag2019MetaLearningWW}
S.~Flennerhag, A.~A. Rusu, R.~Pascanu, F.~Visin, H.~Yin, and R.~Hadsell, ``Meta-learning with warped gradient descent,'' in \emph{ICLR}, 2020.

\bibitem{Shaban2019TruncatedBF}
A.~Shaban, C.~Cheng, N.~Hatch, and B.~Boots, ``Truncated back-propagation for bilevel optimization,'' in \emph{AISTATS}, 2019.

\bibitem{Bertinetto2018MetalearningWD}
L.~Bertinetto, J.~F. Henriques, P.~H.~S. Torr, and A.~Vedaldi, ``Meta-learning with differentiable closed-form solvers,'' in \emph{ICLR}, 2019.

\bibitem{Shin2021LargeScaleMW}
J.~Shin, H.~Lee, B.~Gong, and S.~J. Hwang, ``Large-scale meta-learning with continual trajectory shifting,'' in \emph{ICML}, 2021.

\bibitem{Nichol2018OnFM}
A.~Nichol, J.~Achiam, and J.~Schulman, ``On first-order meta-learning algorithms,'' \emph{arXiv preprint arXiv:1803.02999}, 2018.

\bibitem{Zintgraf2018FastCA}
L.~M. Zintgraf, K.~Shiarlis, V.~Kurin, K.~Hofmann, and S.~Whiteson, ``Fast context adaptation via meta-learning,'' in \emph{ICML}, 2019.

\bibitem{Javed2019MetaLearningRF}
K.~Javed and M.~White, ``Meta-learning representations for continual learning,'' in \emph{NeurIPS}, 2019.

\bibitem{Raghu2019RapidLO}
A.~Raghu, M.~Raghu, S.~Bengio, and O.~Vinyals, ``Rapid learning or feature reuse? towards understanding the effectiveness of {MAML},'' in \emph{ICLR}, 2020.

\bibitem{Ravi2016OptimizationAA}
S.~Ravi and H.~Larochelle, ``Optimization as a model for few-shot learning,'' in \emph{ICLR}, 2017.

\bibitem{Hochreiter2001LearningTL}
S.~Hochreiter, A.~S. Younger, and P.~R. Conwell, ``Learning to learn using gradient descent,'' in \emph{International Conference on Artificial Neural Networks}, 2001.

\bibitem{Mishra2017ASN}
N.~Mishra, M.~Rohaninejad, X.~Chen, and P.~Abbeel, ``A simple neural attentive meta-learner,'' in \emph{ICLR}, 2018.

\bibitem{Santoro2016MetaLearningWM}
A.~Santoro, S.~Bartunov, M.~Botvinick, D.~Wierstra, and T.~P. Lillicrap, ``Meta-learning with memory-augmented neural networks,'' in \emph{ICML}, 2016.

\bibitem{Munkhdalai2017MetaN}
T.~Munkhdalai and H.~Yu, ``Meta networks,'' in \emph{ICML}, 2017.

\bibitem{Koch2015SiameseNN}
G.~Koch, R.~Zemel, R.~Salakhutdinov \emph{et~al.}, ``Siamese neural networks for one-shot image recognition,'' in \emph{ICML deep learning workshop}, 2015.

\bibitem{NIPS2016_90e13578}
O.~Vinyals, C.~Blundell, T.~Lillicrap, K.~Kavukcuoglu, and D.~Wierstra, ``Matching networks for one shot learning,'' in \emph{NeurIPS}, 2016.

\bibitem{Satorras2017FewShotLW}
V.~G. Satorras and J.~B. Estrach, ``Few-shot learning with graph neural networks,'' in \emph{ICLR}, 2018.

\bibitem{Rebuffi2016iCaRL}
S.~Rebuffi, A.~Kolesnikov, G.~Sperl, and C.~H. Lampert, ``icarl: Incremental classifier and representation learning,'' in \emph{CVPR}, 2017.

\bibitem{Rolnick2018ER}
D.~Rolnick, A.~Ahuja, J.~Schwarz, T.~P. Lillicrap, and G.~Wayne, ``Experience replay for continual learning,'' in \emph{NeurIPS}, 2019.

\bibitem{Riemer2018LearningTL}
M.~Riemer, I.~Cases, R.~Ajemian, M.~Liu, I.~Rish, Y.~Tu, and G.~Tesauro, ``Learning to learn without forgetting by maximizing transfer and minimizing interference,'' in \emph{ICLR}, 2019.

\bibitem{Chaudhry2018EfficientLL}
A.~Chaudhry, M.~Ranzato, M.~Rohrbach, and M.~Elhoseiny, ``Efficient lifelong learning with {A-GEM},'' in \emph{ICLR}, 2019.

\bibitem{Aljundi2019GradientBS}
R.~Aljundi, M.~Lin, B.~Goujaud, and Y.~Bengio, ``Gradient based sample selection for online continual learning,'' in \emph{NeurIPS}, 2019.

\bibitem{Shin2017ContinualLW}
H.~Shin, J.~K. Lee, J.~Kim, and J.~Kim, ``Continual learning with deep generative replay,'' in \emph{NeurIPS}, 2017.

\bibitem{Lesort2018GenerativeMF}
T.~Lesort, H.~Caselles-Dupr{\'e}, M.~G. Ortiz, A.~Stoian, and D.~Filliat, ``Generative models from the perspective of continual learning,'' \emph{International Joint Conference on Neural Networks}, 2018.

\bibitem{Sun2019LAMOL}
F.~Sun, C.~Ho, and H.~Lee, ``{LAMOL:} language modeling for lifelong language learning,'' in \emph{ICLR}, 2020.

\bibitem{Wu2018IncrementalCL}
Y.~Wu, Y.~Chen, L.~Wang, Y.~Ye, Z.~Liu, Y.~Guo, Z.~Zhang, and Y.~R. Fu, ``Incremental classifier learning with generative adversarial networks,'' \emph{arXiv preprint arXiv:1802.00853}, 2018.

\bibitem{vandeVen2018GenerativeRW}
G.~M. van~de Ven and A.~S. Tolias, ``Generative replay with feedback connections as a general strategy for continual learning,'' \emph{arXiv preprint arXiv:1809.10635}, 2018.

\bibitem{Yoon2017LifelongLW}
J.~Yoon, E.~Yang, J.~Lee, and S.~J. Hwang, ``Lifelong learning with dynamically expandable networks,'' in \emph{ICLR}, 2018.

\bibitem{Beaulieu2020LearningTC}
S.~L.~E. Beaulieu, L.~Frati, T.~Miconi, J.~Lehman, K.~O. Stanley, J.~Clune, and N.~Cheney, ``Learning to continually learn,'' \emph{arXiv preprint arXiv:2002.09571}, 2020.

\bibitem{AlShedivat2017ContinuousAV}
M.~Al{-}Shedivat, T.~Bansal, Y.~Burda, I.~Sutskever, I.~Mordatch, and P.~Abbeel, ``Continuous adaptation via meta-learning in nonstationary and competitive environments,'' in \emph{ICLR}, 2018.

\bibitem{Kim2020APG}
D.~Kim, M.~Liu, M.~Riemer, C.~Sun, M.~Abdulhai, G.~Habibi, S.~Lopez{-}Cot, G.~Tesauro, and J.~P. How, ``A policy gradient algorithm for learning to learn in multiagent reinforcement learning,'' in \emph{ICML}, 2021.

\bibitem{Banayeeanzade2021GenerativeVD}
M.~Banayeeanzade, R.~Mirzaiezadeh, H.~Hasani, and M.~Soleymani, ``Generative vs. discriminative: Rethinking the meta-continual learning,'' in \emph{NeurIPS}, 2021.

\bibitem{Harrison2018MetaLearningPF}
J.~Harrison, A.~Sharma, and M.~Pavone, ``Meta-learning priors for efficient online bayesian regression,'' in \emph{Workshop on the Algorithmic Foundations of Robotics}, 2018.

\bibitem{Lee2024LearningTC}
S.~Lee, H.~Jeon, J.~Son, and G.~Kim, ``Learning to continually learn with the bayesian principle,'' in \emph{ICML}, 2024.

\bibitem{lee2023recasting}
S.~Lee, J.~Son, and G.~Kim, ``Recasting continual learning as sequence modeling,'' in \emph{NeurIPS}, 2023.

\bibitem{Duan2016RL2FR}
Y.~Duan, J.~Schulman, X.~Chen, P.~L. Bartlett, I.~Sutskever, and P.~Abbeel, ``{RL\textsuperscript{2}: Fast Reinforcement Learning via Slow Reinforcement Learning},'' \emph{arXiv preprint arXiv:1611.02779}, 2016.

\bibitem{Nagabandi2018LearningTA}
A.~Nagabandi, I.~Clavera, S.~Liu, R.~S. Fearing, P.~Abbeel, S.~Levine, and C.~Finn, ``Learning to adapt in dynamic, real-world environments through meta-reinforcement learning,'' in \emph{ICLR}, 2019.

\bibitem{Lu2023StructuredSS}
C.~X. Lu, Y.~Schroecker, A.~Gu, E.~Parisotto, J.~N. Foerster, S.~Singh, and F.~M.~P. Behbahani, ``Structured state space models for in-context reinforcement learning,'' \emph{NeurIPS}, 2023.

\bibitem{Finn2019OnlineM}
C.~Finn, A.~Rajeswaran, S.~M. Kakade, and S.~Levine, ``Online meta-learning,'' in \emph{ICML}, 2019.

\bibitem{Acar2021MemoryEO}
D.~A.~E. Acar, R.~Zhu, and V.~Saligrama, ``Memory efficient online meta learning,'' in \emph{ICML}, 2021.

\bibitem{Yap2020AddressingCF}
P.~C. Yap, H.~Ritter, and D.~Barber, ``Addressing catastrophic forgetting in few-shot problems,'' in \emph{ICML}, 2021.

\bibitem{Caccia2020OnlineFA}
M.~Caccia, P.~Rodr{\'{\i}}guez, O.~Ostapenko, F.~Normandin, M.~Lin, L.~Page{-}Caccia, I.~H. Laradji, I.~Rish, A.~Lacoste, D.~V{\'{a}}zquez, and L.~Charlin, ``Online fast adaptation and knowledge accumulation {(OSAKA):} a new approach to continual learning,'' in \emph{NeurIPS}, 2020.

\bibitem{Clark2022MetaLearningFW}
K.~Clark, K.~Guu, M.-W. Chang, P.~Pasupat, G.~E. Hinton, and M.~Norouzi, ``Meta-learning fast weight language models,'' \emph{EMNLP}, 2022.

\bibitem{Jerfel2018ReconcilingMA}
G.~Jerfel, E.~Grant, T.~Griffiths, and K.~A. Heller, ``Reconciling meta-learning and continual learning with online mixtures of tasks,'' in \emph{NeurIPS}, 2019.

\bibitem{Nagabandi2018DeepOL}
A.~Nagabandi, C.~Finn, and S.~Levine, ``Deep online learning via meta-learning: Continual adaptation for model-based {RL},'' in \emph{ICLR}, 2019.

\bibitem{Zhang2021VariationalCB}
Q.~Zhang, J.~Fang, Z.~Meng, S.~Liang, and E.~Yilmaz, ``Variational continual bayesian meta-learning,'' in \emph{NeurIPS}, 2021.

\bibitem{Yao2020OnlineSM}
H.~Yao, Y.~Zhou, M.~Mahdavi, Z.~Li, R.~Socher, and C.~Xiong, ``Online structured meta-learning,'' in \emph{NeurIPS}, 2020.

\bibitem{pmlr-v202-wu23d}
B.~Wu, J.~Fang, X.~Zeng, S.~Liang, and Q.~Zhang, ``Adaptive compositional continual meta-learning,'' in \emph{ICML}, 2023.

\bibitem{Gupta2020LaMAMLLM}
G.~Gupta, K.~Yadav, and L.~Paull, ``Look-ahead meta learning for continual learning,'' in \emph{NeurIPS}, 2020.

\bibitem{Rajasegaran2020iTAMLAI}
J.~Rajasegaran, S.~Khan, M.~Hayat, F.~S. Khan, and M.~Shah, ``itaml: An incremental task-agnostic meta-learning approach,'' in \emph{CVPR}, 2020.

\bibitem{Volpi2020ContinualAO}
R.~Volpi, D.~Larlus, and G.~Rogez, ``Continual adaptation of visual representations via domain randomization and meta-learning,'' in \emph{CVPR}, 2021.

\bibitem{Joseph2020IncrementalOD}
K.~J. Joseph, J.~Rajasegaran, S.~H. Khan, F.~S. Khan, V.~N. Balasubramanian, and L.~Shao, ``Incremental object detection via meta-learning,'' \emph{IEEE Transactions on Pattern Analysis and Machine Intelligence}, 2020.

\bibitem{Obamuyide2019MetaLearningIL}
A.~Obamuyide and A.~Vlachos, ``Meta-learning improves lifelong relation extraction,'' in \emph{Proceedings of the 4th Workshop on Representation Learning for NLP}, 2019.

\bibitem{Wang2020EfficientML}
Z.~Wang, S.~V. Mehta, B.~Poczos, and J.~Carbonell, ``Efficient meta lifelong-learning with limited memory,'' in \emph{EMNLP}, 2020.

\bibitem{Wu2021CurriculumMetaLF}
T.~Wu, X.~Li, Y.~Li, G.~Haffari, G.~Qi, Y.~Zhu, and G.~Xu, ``Curriculum-meta learning for order-robust continual relation extraction,'' in \emph{AAAI}, 2021.

\bibitem{Jin2021LearnCG}
X.~Jin, B.~Y. Lin, M.~Rostami, and X.~Ren, ``Learn continually, generalize rapidly: Lifelong knowledge accumulation for few-shot learning,'' in \emph{Findings of the Association for Computational Linguistics: EMNLP 2021}, 2021.

\bibitem{wu2024meta}
Y.~Wu, L.-K. Huang, R.~Wang, D.~Meng, and Y.~Wei, ``Meta continual learning revisited: Implicitly enhancing online hessian approximation via variance reduction,'' in \emph{ICLR}, 2024.

\bibitem{Oswald2019ContinualLW}
J.~von Oswald, C.~Henning, J.~Sacramento, and B.~F. Grewe, ``Continual learning with hypernetworks,'' in \emph{ICLR}, 2020.

\bibitem{Hu2018OvercomingCF}
W.~Hu, Z.~Lin, B.~Liu, C.~Tao, Z.~Tao, J.~Ma, D.~Zhao, and R.~Yan, ``Overcoming catastrophic forgetting for continual learning via model adaptation,'' in \emph{ICLR}, 2019.

\bibitem{Joseph2020MetaConsolidationFC}
K.~J. Joseph and V.~N. Balasubramanian, ``Meta-consolidation for continual learning,'' in \emph{NeurIPS}, 2020.

\bibitem{Ehret2020ContinualLI}
B.~Ehret, C.~Henning, M.~R. Cervera, A.~Meulemans, J.~von Oswald, and B.~F. Grewe, ``Continual learning in recurrent neural networks,'' in \emph{ICLR}, 2021.

\bibitem{Chandra2022ContinualLW}
D.~S. Chandra, S.~Varshney, P.~K. Srijith, and S.~Gupta, ``Continual learning with dependency preserving hypernetworks,'' \emph{WACV}, 2022.

\bibitem{Hemati2023PartialHF}
H.~Hemati, V.~Lomonaco, D.~Bacciu, and D.~Borth, ``Partial hypernetworks for continual learning,'' in \emph{CoLLAs}, 2023.

\bibitem{Tao2020FewShotCL}
X.~Tao, X.~Hong, X.~Chang, S.~Dong, X.~Wei, and Y.~Gong, ``Few-shot class-incremental learning,'' in \emph{CVPR}, 2020.

\bibitem{Cheraghian2021SemanticawareKD}
A.~Cheraghian, S.~Rahman, P.~Fang, S.~K. Roy, L.~Petersson, and M.~Harandi, ``Semantic-aware knowledge distillation for few-shot class-incremental learning,'' in \emph{CVPR}, 2021.

\bibitem{Dong2021FewShotCL}
S.~Dong, X.~Hong, X.~Tao, X.~Chang, X.~Wei, and Y.~Gong, ``Few-shot class-incremental learning via relation knowledge distillation,'' in \emph{AAAI}, 2021.

\bibitem{Zhou2022ForwardCF}
D.~Zhou, F.~Wang, H.~Ye, L.~Ma, S.~Pu, and D.~Zhan, ``Forward compatible few-shot class-incremental learning,'' in \emph{CVPR}, 2022.

\bibitem{Cheraghian2021SynthesizedFB}
A.~Cheraghian, S.~Rahman, S.~Ramasinghe, P.~Fang, C.~Simon, L.~Petersson, and M.~Harandi, ``Synthesized feature based few-shot class-incremental learning on a mixture of subspaces,'' in \emph{ICCV}, 2021.

\bibitem{Mazumder2021FewShotLL}
P.~Mazumder, P.~Singh, and P.~Rai, ``Few-shot lifelong learning,'' in \emph{AAAI}, 2021.

\bibitem{Hersche2022ConstrainedFC}
M.~Hersche, G.~Karunaratne, G.~Cherubini, L.~Benini, A.~Sebastian, and A.~Rahimi, ``Constrained few-shot class-incremental learning,'' in \emph{CVPR}, 2022.

\bibitem{Zhu2021SelfPromotedPR}
K.~Zhu, Y.~Cao, W.~Zhai, J.~Cheng, and Z.~Zha, ``Self-promoted prototype refinement for few-shot class-incremental learning,'' in \emph{CVPR}, 2021.

\bibitem{Zhang2021FewShotIL}
C.~Zhang, N.~Song, G.~Lin, Y.~Zheng, P.~Pan, and Y.~Xu, ``Few-shot incremental learning with continually evolved classifiers,'' in \emph{CVPR}, 2021, pp. 12\,455--12\,464.

\bibitem{Zhu2023FewshotIL}
Z.~Zhu, P.~Wang, W.~Diao, J.~Yang, H.~Wang, and X.~Sun, ``Few-shot incremental learning with continual prototype calibration for remote sensing image fine-grained classification,'' \emph{ISPRS Journal of Photogrammetry and Remote Sensing}, 2023.

\bibitem{Zhao2023FewShotCL}
L.~Zhao, J.~Lu, Y.~Xu, Z.~Cheng, D.~Guo, Y.~Niu, and X.~Fang, ``Few-shot class-incremental learning via class-aware bilateral distillation,'' in \emph{CVPR}, 2023.

\bibitem{Zhuang2023GKEALGK}
H.~Zhuang, Z.~Weng, R.~He, Z.~Lin, and Z.~Zeng, ``Gkeal: Gaussian kernel embedded analytic learning for few-shot class incremental task,'' \emph{CVPR}, 2023.

\bibitem{Chi2022MetaFSCILAM}
Z.~Chi, L.~Gu, H.~Liu, Y.~Wang, Y.~Yu, and J.~Tang, ``Metafscil: {A} meta-learning approach for few-shot class incremental learning,'' in \emph{CVPR}, 2022.

\bibitem{Zhou2022FewShotCL}
D.-W. Zhou, H.-J. Ye, L.~Ma, D.~Xie, S.~Pu, and D.-C. Zhan, ``Few-shot class-incremental learning by sampling multi-phase tasks,'' \emph{IEEE Transactions on Pattern Analysis and Machine Intelligence}, 2022.

\bibitem{PrezRa2020IncrementalFO}
J.~P{\'{e}}rez{-}R{\'{u}}a, X.~Zhu, T.~M. Hospedales, and T.~Xiang, ``Incremental few-shot object detection,'' in \emph{CVPR}, 2020.

\bibitem{Li2020IncrementalFO}
Y.~Li, H.~Zhu, S.~Tian, F.~Feng, J.~Ma, C.~S. Teo, C.~Xiang, P.~Vadakkepat, and T.~H. Lee, ``Incremental few-shot object detection for robotics,'' \emph{ICRA}, 2020.

\bibitem{Dong2022IncrementalDETRIF}
N.~Dong, Y.~Zhang, M.~Ding, and G.~H. Lee, ``Incremental-detr: Incremental few-shot object detection via self-supervised learning,'' in \emph{AAAI}, 2023.

\bibitem{Cheng2022MetaLearningBasedIF}
M.~Cheng, H.~Wang, and Y.~Long, ``Meta-learning-based incremental few-shot object detection,'' \emph{IEEE Transactions on Circuits and Systems for Video Technology}, 2022.

\bibitem{Yin2022SylphAH}
L.~Yin, J.-M. P{\'e}rez-R{\'u}a, and K.~J. Liang, ``Sylph: A hypernetwork framework for incremental few-shot object detection,'' \emph{CVPR}, 2022.

\bibitem{Choi2023IncrementalFO}
T.~Choi and J.-H. Kim, ``Incremental few-shot object detection via simple fine-tuning approach,'' \emph{ICRA}, 2023.

\bibitem{Finn2018Universality}
C.~Finn and S.~Levine, ``Meta-learning and universality: Deep representations and gradient descent can approximate any learning algorithm,'' in \emph{ICLR}, 2018.

\bibitem{Chaudhry2018RiemannianWF}
A.~Chaudhry, P.~K. Dokania, T.~Ajanthan, and P.~H.~S. Torr, ``Riemannian walk for incremental learning: Understanding forgetting and intransigence,'' in \emph{ECCV}, 2018.

\bibitem{Nguyen2017VariationalCL}
C.~V. Nguyen, Y.~Li, T.~D. Bui, and R.~E. Turner, ``Variational continual learning,'' in \emph{ICLR}, 2018.

\bibitem{Farquahr2019Bayesian}
S.~Farquhar and Y.~Gal, ``A unifying bayesian view of continual learning,'' \emph{arXiv preprint arXiv:1902.06494}, 2019.

\bibitem{Fisher1934}
R.~A. Fisher, ``Two new properties of mathematical likelihood,'' \emph{Proceedings of the Royal Society of London. Series A, Containing Papers of a Mathematical and Physical Character}, 1934.

\bibitem{Darmois1935}
G.~Darmois, ``Sur les lois de probabilit{\'e}a estimation exhaustive,'' \emph{CR Acad. Sci. Paris}, 1935.

\bibitem{Pitman1936}
E.~J.~G. Pitman, ``Sufficient statistics and intrinsic accuracy,'' in \emph{Mathematical Proceedings of the Cambridge Philosophical Society}, 1936.

\bibitem{Koopman1936}
B.~O. Koopman, ``On distributions admitting a sufficient statistic,'' \emph{Transactions of the American Mathematical society}, 1936.

\bibitem{Brown2020LanguageMA}
T.~B. Brown, B.~Mann, N.~Ryder, M.~Subbiah, J.~Kaplan, P.~Dhariwal, A.~Neelakantan, P.~Shyam, G.~Sastry, A.~Askell, S.~Agarwal, A.~Herbert{-}Voss, G.~Krueger, T.~Henighan, R.~Child, A.~Ramesh, D.~M. Ziegler, J.~Wu, C.~Winter, C.~Hesse, M.~Chen, E.~Sigler, M.~Litwin, S.~Gray, B.~Chess, J.~Clark, C.~Berner, S.~McCandlish, A.~Radford, I.~Sutskever, and D.~Amodei, ``Language models are few-shot learners,'' in \emph{NeurIPS}, 2020.

\bibitem{Vaswani2017AttentionIA}
A.~Vaswani, N.~Shazeer, N.~Parmar, J.~Uszkoreit, L.~Jones, A.~N. Gomez, L.~Kaiser, and I.~Polosukhin, ``Attention is all you need,'' in \emph{NeurIPS}, 2017.

\bibitem{Katharopoulos2020TransformersAR}
A.~Katharopoulos, A.~Vyas, N.~Pappas, and F.~Fleuret, ``Transformers are rnns: Fast autoregressive transformers with linear attention,'' in \emph{ICML}, 2020.

\bibitem{Choromanski2020RethinkingAW}
K.~M. Choromanski, V.~Likhosherstov, D.~Dohan, X.~Song, A.~Gane, T.~Sarl{\'{o}}s, P.~Hawkins, J.~Q. Davis, A.~Mohiuddin, L.~Kaiser, D.~B. Belanger, L.~J. Colwell, and A.~Weller, ``Rethinking attention with performers,'' in \emph{ICLR}, 2021.

\bibitem{Tay20M23Efficient}
Y.~Tay, M.~Dehghani, D.~Bahri, and D.~Metzler, ``Efficient transformers: A survey,'' \emph{ACM Computing Surveys}, 2020.

\bibitem{CurriculumLearning}
Y.~Bengio, J.~Louradour, R.~Collobert, and J.~Weston, ``Curriculum learning,'' in \emph{ICML}, 2009.

\bibitem{Sutton2018Reinforcement}
R.~S. Sutton and A.~G. Barto, \emph{Reinforcement Learning: An Introduction}.\hskip 1em plus 0.5em minus 0.4em\relax MIT press, 2018.

\bibitem{TFLength}
C.~Anil, Y.~Wu, A.~Andreassen, A.~Lewkowycz, V.~Misra, V.~V. Ramasesh, A.~Slone, G.~Gur{-}Ari, E.~Dyer, and B.~Neyshabur, ``Exploring length generalization in large language models,'' in \emph{NeurIPS}, 2022.

\bibitem{ALiBi}
O.~Press, N.~A. Smith, and M.~Lewis, ``Train short, test long: Attention with linear biases enables input length extrapolation,'' in \emph{ICLR}, 2022.

\bibitem{ruoss-2023-randomized}
A.~Ruoss, G.~Del{\'e}tang, T.~Genewein, J.~Grau-Moya, R.~Csord{\'a}s, M.~Bennani, S.~Legg, and J.~Veness, ``Randomized positional encodings boost length generalization of {Transformers},'' in \emph{ACL}, 2023.

\bibitem{Denevi2019OnlineWithinOnlineM}
G.~Denevi, D.~Stamos, C.~Ciliberto, and M.~Pontil, ``Online-within-online meta-learning,'' in \emph{NeurIPS 2019}, 2019.

\bibitem{Harrison2019ContinuousMW}
J.~Harrison, A.~Sharma, C.~Finn, and M.~Pavone, ``Continuous meta-learning without tasks,'' in \emph{NeurIPS}, 2020.

\bibitem{Ren2020WanderingWA}
M.~Ren, M.~L. Iuzzolino, M.~C. Mozer, and R.~S. Zemel, ``Wandering within a world: Online contextualized few-shot learning,'' in \emph{ICLR}, 2021.

\bibitem{Hannan1957Approximation}
J.~Hannan, ``Approximation to bayes risk in repeated play,'' \emph{Contributions to the Theory of Games}, 1957.

\bibitem{Kalai2005Efficient}
A.~T. Kalai and S.~S. Vempala, ``Efficient algorithms for online decision problems,'' \emph{Journal of Computer and System Sciences}, 2005.

\bibitem{871c6579c7e14fa19571ab1b4e6f6f22}
M.~Opper and O.~Winther, \emph{A Bayesian approach to on-line learning}, D.~Saad, Ed.\hskip 1em plus 0.5em minus 0.4em\relax Cambridge University Press, 1999.

\bibitem{Ritter2018OnlineSL}
H.~Ritter, A.~Botev, and D.~Barber, ``Online structured laplace approximations for overcoming catastrophic forgetting,'' in \emph{NeurIPS}, 2018.

\bibitem{McLachlan2000Mixture}
G.~J. McLachlan, S.~X. Lee, and S.~I. Rathnayake, ``Finite mixture models,'' \emph{Annual Review of Statistics and Its Application}, 2019.

\bibitem{Antoniak1974MixturesOD}
C.~E. Antoniak, ``Mixtures of dirichlet processes with applications to bayesian nonparametric problems,'' \emph{Annals of Statistics}, 1974.

\bibitem{FERGUSON1983287}
T.~S. Ferguson, ``Bayesian density estimation by mixtures of normal distributions,'' in \emph{Recent Advances in Statistics}, M.~H. Rizvi, J.~S. Rustagi, and D.~Siegmund, Eds.\hskip 1em plus 0.5em minus 0.4em\relax Academic Press, 1983.

\bibitem{Lin2013OnlineLO}
D.~Lin, ``Online learning of nonparametric mixture models via sequential variational approximation,'' in \emph{NeurIPS}, 2013.

\bibitem{Hjort1990Beta}
N.~L. Hjort, ``Nonparametric bayes estimators based on beta processes in models for life history data,'' \emph{the Annals of Statistics}, 1990.

\bibitem{Griffiths2011TheIB}
T.~L. Griffiths and Z.~Ghahramani, ``The indian buffet process: An introduction and review.'' \emph{Journal of Machine Learning Research}, 2011.

\bibitem{Meeds2006Factor}
E.~Meeds, Z.~Ghahramani, R.~M. Neal, and S.~T. Roweis, ``Modeling dyadic data with binary latent factors,'' in \emph{NeurIPS}, 2006.

\bibitem{Oswald2021LearningWT}
J.~von Oswald, D.~Zhao, S.~Kobayashi, S.~Schug, M.~Caccia, N.~Zucchet, and J.~Sacramento, ``Learning where to learn: Gradient sparsity in meta and continual learning,'' in \emph{NeurIPS}, 2021.

\bibitem{Netzer2011ReadingDI}
Y.~Netzer, T.~Wang, A.~Coates, A.~Bissacco, B.~Wu, and A.~Y. Ng, ``Reading digits in natural images with unsupervised feature learning,'' in \emph{NIPS Workshop on Deep Learning and Unsupervised Feature Learning}, 2011.

\bibitem{Ganin2014UnsupervisedDA}
Y.~Ganin and V.~S. Lempitsky, ``Unsupervised domain adaptation by backpropagation,'' in \emph{ICML}, 2015.

\bibitem{Ha2016HyperNetworks}
D.~Ha, A.~M. Dai, and Q.~V. Le, ``Hypernetworks,'' in \emph{ICLR}, 2017.

\bibitem{Jia2016DynamicFN}
X.~Jia, B.~D. Brabandere, T.~Tuytelaars, and L.~V. Gool, ``Dynamic filter networks,'' in \emph{NeurIPS}, 2016.

\bibitem{Krueger2017BayesianH}
D.~Krueger, C.-W. Huang, R.~Islam, R.~Turner, A.~Lacoste, and A.~C. Courville, ``Bayesian hypernetworks,'' \emph{arXiv preprint arXiv:1710.04759}, 2017.

\bibitem{Savarese2019LearningIR}
P.~Savarese and M.~Maire, ``Learning implicitly recurrent cnns through parameter sharing,'' in \emph{ICLR}, 2019.

\bibitem{Bertinetto2016LearningFO}
L.~Bertinetto, J.~F. Henriques, J.~Valmadre, P.~H.~S. Torr, and A.~Vedaldi, ``Learning feed-forward one-shot learners,'' in \emph{NeurIPS}, 2016.

\bibitem{Zhao2020MetaLearningVH}
D.~Zhao, S.~Kobayashi, J.~Sacramento, and J.~von Oswald, ``Meta-learning via hypernetworks,'' in \emph{4th Workshop on Meta-Learning at NeurIPS}, 2020.

\bibitem{Zhmoginov2022HyperTransformerMG}
A.~Zhmoginov, M.~Sandler, and M.~Vladymyrov, ``Hypertransformer: Model generation for supervised and semi-supervised few-shot learning,'' in \emph{ICML}, 2022.

\bibitem{Tolstikhin2017WassersteinA}
I.~O. Tolstikhin, O.~Bousquet, S.~Gelly, and B.~Sch{\"{o}}lkopf, ``Wasserstein auto-encoders,'' in \emph{ICLR}, 2018.

\bibitem{Shmelkov2017IncrementalLO}
K.~Shmelkov, C.~Schmid, and K.~Alahari, ``Incremental learning of object detectors without catastrophic forgetting,'' in \emph{ICCV}, 2017.

\bibitem{Feng2022OvercomingCF}
T.~Feng, M.~Wang, and H.~Yuan, ``Overcoming catastrophic forgetting in incremental object detection via elastic response distillation,'' in \emph{CVPR}, 2022.

\bibitem{Zhou2019ObjectsAP}
X.~Zhou, D.~Wang, and P.~Kr{\"a}henb{\"u}hl, ``Objects as points,'' \emph{arXiv preprint arXiv:1904.07850}, 2019.

\bibitem{Tian2019FCOSFC}
Z.~Tian, C.~Shen, H.~Chen, and T.~He, ``{FCOS:} fully convolutional one-stage object detection,'' in \emph{ICCV}, 2019.

\bibitem{Qiao2017FewShotIR}
S.~Qiao, C.~Liu, W.~Shen, and A.~L. Yuille, ``Few-shot image recognition by predicting parameters from activations,'' in \emph{CVPR}, 2018.

\bibitem{Gidaris2018DynamicFV}
S.~Gidaris and N.~Komodakis, ``Dynamic few-shot visual learning without forgetting,'' in \emph{CVPR}, 2018.

\bibitem{Ren2018IncrementalFL}
M.~Ren, R.~Liao, E.~Fetaya, and R.~S. Zemel, ``Incremental few-shot learning with attention attractor networks,'' in \emph{NeurIPS}, 2019.

\bibitem{Bing2022MetaReinforcementLI}
Z.~Bing, D.~Lerch, K.~Huang, and A.~Knoll, ``Meta-reinforcement learning in non-stationary and dynamic environments,'' \emph{IEEE Transactions on Pattern Analysis and Machine Intelligence}, 2022.

\bibitem{Gu2021EfficientlyML}
A.~Gu, K.~Goel, and C.~R'e, ``Efficiently modeling long sequences with structured state spaces,'' \emph{ICLR}, 2021.

\bibitem{Lazaridou2021MindTG}
A.~Lazaridou, A.~Kuncoro, E.~Gribovskaya, D.~Agrawal, A.~Liska, T.~Terzi, M.~Gimenez, C.~de~Masson~d'Autume, T.~Kocisk{\'y}, S.~Ruder, D.~Yogatama, K.~Cao, S.~Young, and P.~Blunsom, ``Mind the gap: Assessing temporal generalization in neural language models,'' in \emph{NeurIPS}, 2021.

\bibitem{Jang2021TowardsCK}
J.~Jang, S.~Ye, S.~Yang, J.~Shin, J.~Han, G.~Kim, S.~J. Choi, and M.~Seo, ``Towards continual knowledge learning of language models,'' \emph{ICLR}, 2021.

\bibitem{Jang2022TemporalWikiAL}
J.~Jang, S.~Ye, C.~Lee, S.~Yang, J.~Shin, J.~Han, G.~Kim, and M.~Seo, ``Temporalwiki: A lifelong benchmark for training and evaluating ever-evolving language models,'' in \emph{EMNLP}, 2022.

\bibitem{Livska2022StreamingQAAB}
A.~Livska, T.~Kovcisk'y, E.~Gribovskaya, T.~Terzi, E.~Sezener, D.~Agrawal, C.~de~Masson~d'Autume, T.~Scholtes, M.~Zaheer, S.~Young, E.~Gilsenan-McMahon, S.~Austin, P.~Blunsom, and A.~Lazaridou, ``Streamingqa: A benchmark for adaptation to new knowledge over time in question answering models,'' \emph{ICML}, 2022.

\bibitem{Kim2023CarpeDO}
Y.~Kim, J.~Yoon, S.~Ye, S.~J. Hwang, and S.~young Yun, ``Carpe diem: On the evaluation of world knowledge in lifelong language models,'' \emph{NAACL}, 2023.

\bibitem{Dhingra2021TimeAwareLM}
B.~Dhingra, J.~R. Cole, J.~M. Eisenschlos, D.~Gillick, J.~Eisenstein, and W.~W. Cohen, ``Time-aware language models as temporal knowledge bases,'' \emph{Transactions of the Association for Computational Linguistics}, 2021.

\bibitem{Hu2023MetaLearningOA}
N.~J. Hu, E.~Mitchell, C.~D. Manning, and C.~Finn, ``Meta-learning online adaptation of language models,'' \emph{EMNLP}, 2023.

\bibitem{ImitationLearningSurvey}
M.~Zare, P.~M. Kebria, A.~Khosravi, and S.~Nahavandi, ``A survey of imitation learning: Algorithms, recent developments, and challenges,'' \emph{CoRR}, 2023.

\bibitem{ActiveLearningSurvey}
B.~Settles, ``Active learning literature survey,'' University of Wisconsin-Madison Department of Computer Sciences, Tech. Rep., 2009.

\bibitem{RoPE}
J.~Su, M.~H.~M. Ahmed, Y.~Lu, S.~Pan, W.~Bo, and Y.~Liu, ``{RoFormer}: Enhanced {Transformer} with rotary position embedding,'' \emph{Neurocomputing}, 2024.

\bibitem{LengthExtrapolationPE}
L.~Zhao, X.~Feng, X.~Feng, B.~Qin, and T.~Liu, ``Length extrapolation of transformers: {A} survey from the perspective of position encoding,'' \emph{CoRR}, 2023.

\bibitem{wang2023comprehensive}
L.~Wang, X.~Zhang, H.~Su, and J.~Zhu, ``A comprehensive survey of continual learning: Theory, method and application,'' \emph{IEEE Transactions on Pattern Analysis and Machine Intelligence}, 2024.

\end{thebibliography}

\vfill
\vspace{\stretch{5}}

\begin{IEEEbiography}
[{\includegraphics[width=1in,height=1.25in,clip,keepaspectratio]{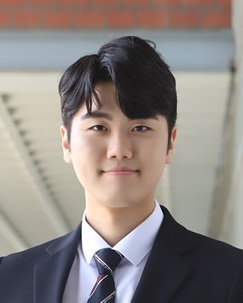}}]{Jaehyeon Son}
  is a student researcher at Seoul National University, supervised by Gunhee Kim.
  He received his BS degree in Statistics and Artificial Intelligence from Seoul National University.
  His research focuses on addressing reinforcement learning and continual learning with the Transformer's transfer learning ability.
\end{IEEEbiography}

\begin{IEEEbiography}[{\includegraphics[width=1in,height=1.25in,clip,keepaspectratio]{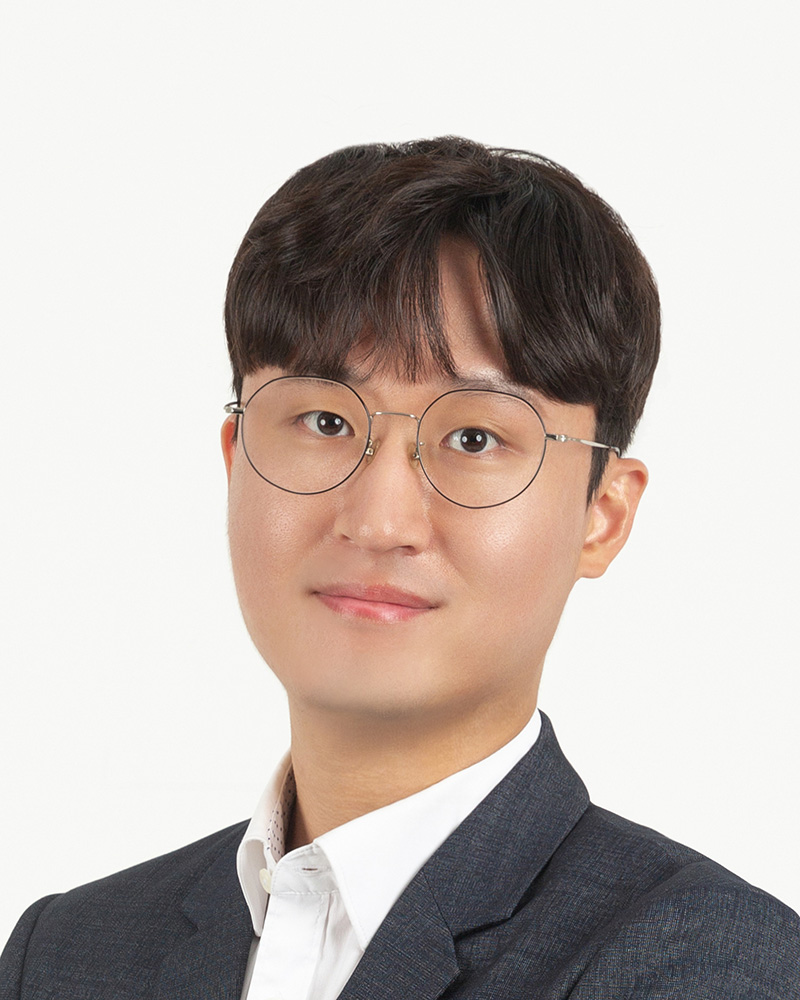}}]{Soochan Lee}
  is a research scientist at LG AI Research.
  He received his BS, MS, and PhD degrees from the Department of Computer Science and Engineering at Seoul National University, where he was awarded the Best Thesis Award for his PhD.
  His research interest includes continual learning, meta-learning, and sequence modeling.
\end{IEEEbiography}

\begin{IEEEbiography}[{\includegraphics[width=1in,height=1.25in,clip,keepaspectratio]{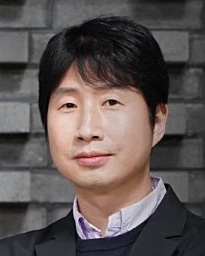}}]{Gunhee Kim}
  is a professor in the Department of Computer Science and Engineering of Seoul National University from 2015. He was a postdoctoral researcher at Disney Research for one and a half years. He received his PhD in 2013 from Computer Science Department of Carnegie Mellon University (CMU). Prior to starting PhD study in 2009, he earned a master’s degree in Robotics Institute of CMU. His research interests are solving computer vision and natural language problems that emerge from big multimodal data shared online. He is a recipient of 2014 ACM SIGKDD doctoral dissertation award, 2015 Naver new faculty award, Best full paper runner-up at ACM VRST 2019, and Outstanding paper award at EMNLP 2023.
\end{IEEEbiography}

\vfill

\end{document}